\documentclass[journal]{IEEEtran}
\usepackage {multirow}
\usepackage{color}
\usepackage[colorlinks,linkcolor=black]{hyperref}
\hypersetup{urlcolor=black}
\usepackage{amsmath}
\ifCLASSINFOpdf
   \usepackage[pdftex]{graphicx}
   \graphicspath{{../pdf/}{../jpeg/}}
   \DeclareGraphicsExtensions{.pdf,.jpeg,.png}
\else
   \usepackage[dvips]{graphicx}
   \graphicspath{{../eps/}}
   \DeclareGraphicsExtensions{.eps}
\fi
\begin{document}
%
% paper title
% Titles are generally capitalized except for words such as a, an, and, as,
% at, but, by, for, in, nor, of, on, or, the, to and up, which are usually
% not capitalized unless they are the first or last word of the title.
% Linebreaks \\ can be used within to get better formatting as desired.
% Do not put math or special symbols in the title.
\title{SwinFuse: A Residual Swin Transformer Fusion Network for Infrared and Visible Images}

%
%
% author names and IEEE memberships
% note positions of commas and nonbreaking spaces ( ~ ) LaTeX will not break
% a structure at a ~ so this keeps an author's name from being broken across
% two lines.
% use \thanks{} to gain access to the first footnote area
% a separate \thanks must be used for each paragraph as LaTeX2e's \thanks
% was not built to handle multiple paragraphs
%

\author{Zhishe~Wang,\textit{~Member,~IEEE},~Yanlin~Chen,~Wenyu~Shao,~Hui~Li,~Lei~Zhang
% <-this % stops a space
\thanks{This work is supported in part by the Fundamental Research Program of Shanxi Province under Grant 201901D111260, in part by the Open Foundation of Shanxi Key Laboratory of Signal Capturing \& Processing under Grant ISPT2020-4. (\textsl{Corresponding author: Zhishe~Wang}).}% <-this % stops a space
\thanks{Zhishe~Wang, Yanlin~Chen and Wenyu~Shao are with the School of Applied Science, Taiyuan University of Science and Technology, Taiyuan, 030024, China. (e-mail: wangzs@tyust.edu.cn; chentyust@163.com; wyshaotyust@163.com).}

\thanks{Hui~Li is with the School of Artificial Intelligence and Computer Science, Jiangnan University, Wuxi, 214122, China (e-mail: lihui.cv@jiangnan.edu.cn).}

\thanks{Lei~Zhang is with the School of Mechatronic Engineering, Nanyang Normal University, Nanyang, 473061, China (e-mail: zhanglei000223@163.com).}}

\maketitle

% As a general rule, do not put math, special symbols or citations
% in the abstract or keywords.
\begin{abstract}
The existing deep learning fusion methods mainly concentrate on the convolutional neural networks, and few attempts are made with transformer. Meanwhile, the convolutional operation is a content-independent interaction between the image and convolution kernel, which may lose some important contexts and further limit fusion performance. Towards this end, we present a simple and strong fusion baseline for infrared and visible images, namely\textit{ Residual Swin Transformer Fusion Network}, termed as SwinFuse. Our SwinFuse includes three parts: the global feature extraction, fusion layer and feature reconstruction. In particular, we build a fully attentional feature encoding backbone to model the long-range dependency, which is a pure transformer network and has a stronger representation ability compared with the convolutional neural networks. Moreover, we design a novel feature fusion strategy based on $L_{1}$-norm for sequence matrices, and measure the corresponding activity levels from row and column vector dimensions, which can well retain competitive infrared brightness and distinct visible details. Finally, we testify our SwinFuse with nine state-of-the-art traditional and deep learning methods on three different datasets through subjective observations and objective comparisons, and the experimental results manifest that the proposed SwinFuse obtains surprising fusion performance with strong generalization ability and competitive computational efficiency. The code will be available at \textit{\url{https://github.com/Zhishe-Wang/SwinFuse}}.

\end{abstract}

% Note that keywords are not normally used for peerreview papers.

% For peer review papers, you can put extra information on the cover
% page as needed:
% \ifCLASSOPTIONpeerreview
% \begin{center} \bfseries EDICS Category: 3-BBND \end{center}
% \fi
%
% For peerreview papers, this IEEEtran command inserts a page break and
% creates the second title. It will be ignored for other modes.
\IEEEpeerreviewmaketitle

\section{Introduction}
% The very first letter is a 2 line initial drop letter followed
% by the rest of the first word in caps.
% 
% form to use if the first word consists of a single letter:
% \IEEEPARstart{A}{demo} file is ....
% 
% form to use if you need the single drop letter followed by
% normal text (unknown if ever used by the IEEE):
% \IEEEPARstart{A}{}demo file is ....
% 
% Some journals put the first two words in caps:
% \IEEEPARstart{T}{his demo} file is ....
% 
% Here we have the typical use of a "T" for an initial drop letter
% and "HIS" in caps to complete the first word.

\IEEEPARstart{I}{nfrared} sensor can detect the hidden or camouflage targets by capturing the thermal radiation energy, and has the strong anti-interference ability with all-time and all-weather, but it cannot acquire typical background details and structural texture. On the contrary, visible sensor can perceive the scene details, color information and texture characteristics by receiving the reflected light, but it fails to distinguish the prominent targets, and is easily affected by the weather and light variations. Considering their complementarity of imaging mechanisms and working conditions, image fusion technology aims to transform their complementary features into a synthesized image through a specific algorithm. The obtained fusion image is more in line with human visual perception, and for other subsequent computer visual tasks, such as object detection and object recognition, which can achieve more accurate decisions than a single sensor. Therefore, infrared and visible image fusion cooperates with these two sensors to generate a higher quality result, and has important applications in many fields, such as person re-recognition [1], object fusion tracking [2] and salient object detection [3] and so on.  

Generally, the core problems of infrared and visible image fusion are how to effectively extract and combine their complementary features. The traditional methods, such as multi-scale transform [4], sparse representation [5], hybrid methods [6], saliency-based [7, 8] and others [9, 10], usually design a fixed representation model to extract features, adopt a specific fusion strategy for their corresponding combination, and then reconstruct a final result by the inverse operations. For example, Li \textit{et al.} introduced MDLatLRR [5] where the latent low-rank representation was designed for the base and detail feature model, the weighted averaging and nuclear norm were proposed as fusion strategies. However, different imaging mechanisms represent contrasting modal characteristics, infrared image perceives the prominent targets through pixel brightness, while visible image describes the structural texture details through gradients and edges. The traditional methods do not fully take into account their modal discrepancies, and adopt the same representation model to distinctively extract features, which may fail to acquire the most effective inherent features, and is easy to produce a weak fusion performance. Besides, the designed fusion rules are usually hand-crafted and intends to become more and more complex, which inevitably limits the relevant practical applications.

\begin{figure*}[!t]
	\centering
	\includegraphics[width=1\textwidth]{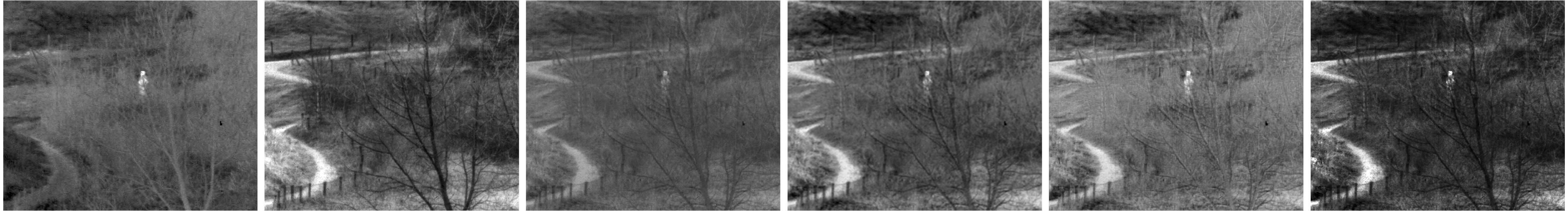}
	% where an .eps filename suffix will be assumed under latex, 
	% and a .pdf suffix will be assumed for pdflatex; or what has been declared
	% via \DeclareGraphicsExtensions.
	\caption{A contrastive example of \textit{Sandpath} selected from TNO dataset. These images, in sequence, are infrared and visible images, the results of MDLatLRR [5], DenseFuse [12], FusionGAN [13] and our SwinFuse. Our result has more prominent target perception and more clarity detail description.}
	\label{Fig1}
\end{figure*}

Different from traditional methods, deep learning can automatically extract different modal features by using a series of learnable filter banks, and has the strong nonlinear fitting ability to establish the complicated relationship between the input and output [11]. Typically, Li \textit{et al.} presented DenseFuse [12], which proposed a convolutional neural network (CNN) to replace the traditional representation model for better feature extraction and reconstruction, and manually designed the corresponding fusion strategies. Ma \textit{et al.} introduced FusionGAN [13] in which the generator was continually optimized by the discriminator with adversarial learning to generate the desired fusion image, avoiding the manual design of activity level and fusion rules. Although their methods have achieved amazing performance, some issues need to be further addressed. Notably, the quality of the obtained fusion image is not only related to pixels within a small perceptive field, but also the pixel intensity and texture details of the overall image. On the one hand, their methods are constrained by the basic convolutional principle, and the interaction between the image and convolution kernel is content-independent. Utilizing the uniform convolution kernel to extract features from disparate modal images may not be the most effective way. On the another hand, their methods are based on the guideline of local processing, and establish the local deep features through a limited receptive field, but cannot model their long-range dependencies, which may lose some important contexts.

To address the above issues, we present a simple and strong baseline based on Swin Transformer [14] for infrared and visible image fusion, namely SwinFuse. Swin Transformer firstly partitions non-overlapping windows to model local attention, and then periodically shifts these windows to bridge global attention. More particularly, we propose residual Swin Transformer blocks, which are composed of several Swin Transformer layers, as backbone to extract the global features. Meanwhile, we adopt the residual connection as a shortcut to implement low-level feature aggregation and information retention. Our SwinFuse makes use of the fully attentional model to interact with image content and attention weights, and has a powerful ability for the long-range dependencies modeling, which can unleash the limitation of existing deep learning based models, and significantly promote the infrared and visible image fusion performance to a new level.

To demonstrate the visual performance of our SwinFuse, Fig.1 gives a contrastive example of \textit{Sandpath} selected from the TNO dataset [15]. From the intuitive observation, MDLatLRR [5] and DenseFuse [12] preserve abundant texture details from visible image, but lose the typical target information from infrared image. On the contrary, FusionGAN [13] retains the high-brightness target of infrared image, while the corresponding target edges are fairly obscure and texture details are seriously losing. However, our SwinFuse achieves satisfactory results in reserving prominent infrared targets and rich visible details, and has a better image contrast.

Our SwinFuse includes three main contributions:

$ \bullet $ We build a fully attentional feature encoding backbone to model the long-range dependency, which only applies the pure transformer without convolutional neural networks. The obtained global attention features have the stronger representation ability in focusing on infrared target perception and visible detail description.

$ \bullet $ We design a novel feature fusion strategy based on $L_{1}$-norm for sequence matrices. The activity levels of source images are measured from row and column vector dimensions, respectively. With this strategy, the obtained results can well retain the competitive brightness of infrared targets and distinct details of visible background.

$ \bullet $ We propose an infrared and visible image fusion transformer framework, and conduct a mass of experiments on different testing datasets. Our SwinFuse achieves surprising results and generalization, which transcends other state-of-the-art deep learning based methods in terms of subjective observation and objective comparison.

The rest of this paper is arranged as follows. Section II mainly introduces transformer in vision tasks and deep learning based methods. Section III illustrates the proposed network architecture and design philosophy. The experiments and discussions are presented in Section IV, and Section V draws the relevant conclusions.

\section{Related Work}

In this section, we firstly introduce the application of the transformer in some vision tasks, and then emphatically illustrate the development of deep learning based methods.

\subsection{Transformer in Vision Tasks}

Transformer [16] was originally designed for machine translation, and had achieved great success in natural language processing (NLP). In 2020, Dosovitskiy \textit{et al.} presented Vision Transformer (ViT) [17], which divided an image into 16$ \times $16 patches, and directly feed these patches into the standard transformer encoder. ViT splits an image into a linear embedded sequence and models the long-range dependency with self-attention meschanism, which generates promising results on certain tasks, such as image classification [18] and image retrieval [19]. However, high resolution images and visual elements varying in scale are very different from word tokens of NLP, which brings some significant challenges for adapting transformer from language domain to vision domain, specially transformer performance and computational efficiency.

To overcome the above limitations, Liu \textit{et al.} developed Swin Transformer [14] in which images were partitioned into local windows and cross-window through shifted operation, and limited attentional calculation in a corresponding window. Thus, its hierarchical architecture introduced the locality of convolutional operation, and obtained a lower computational complexity that is linear with the image size. Inspired by this work, researchers investigated its superiority for other computer vision tasks. For example, Liang \textit{et al.} presented SwinIR [20] for image restoration, which first proposed the convolutional layer to extract shallow features, and then adopted Swin transformer for deep feature extraction. Lin \textit{et al.} introduced SwinTrack [21] to interact with the target object and search region for tracking. However, few studies have developed transformer into image fusion fields.

\subsection{Deep Learning-based Fusion Methods}

Recently, deep learning models [22-31] possess strong capacities in terms of feature extraction and nonlinear data fitting, which have become the mainstream direction of image fusion tasks. Typically, Li \textit{et al.} introduced DenseFuse [12] where an encoder with a convolution layer and a dense block was proposed for feature extraction, and a decoder including four convolution layers was used for feature reconstruction. Zhang \textit{et al.} presented IFCNN [24] where the encoder and decoder respectively included two convolution layers, elementwise-maximum, minimum and mean were applied for fusion rules. These methods design the simple network and propose some appropriate fusion rules, but fail to consider the long-range dependency. To achieve better fusion performance, Jian \textit{et al.} proposed a symmetric feature encoding and decoding network, namely SEDRFuse [25], which applied feature compensation and attention fusion to improve fusion performance. Wang \textit{et al.} presented Res2Fusion [26] where multi-fields aggregated feature encoding backbone was constructed, and double nonlocal attention models were used for fusion strategies. Meanwhile, they also introduced UNFusion [27] in which a densely connected feature encoding and decoding network was exploited, and normalized attention models were designed to model the global dependency. The above methods need to manually design the corresponding fusion rules, and propose a non-end-to-end fusion model. To address this issue, Li \textit{et al.} developed a two-stage learnable network, termed as RFN-Nest [28], which was an improved version of NestFuse [29], and proposed a learnable residual fusion network to replace hand-craft fusion strategies. Furthermore, PMGI [30] and U2Fusion [31] proposed a unified end-to-end network to satisfy several different fusion tasks simultaneously.

In addition, some researchers had exploited the generative adversarial network (GAN) [32-36] for image fusion, and achieved satisfactory results to some extent. Typically, Ma \textit{et al.} firstly presented FusionGAN [13] where the adversarial learning network including a generator and a discriminator was proposed. Since only a discriminator is used, their results are biased towards infrared images and lack visible texture details. Subsequently, they also exploited dual-discriminator architecture, namely DDcGAN [32], to overcome the lack of a single discriminator and apply it for multi-resolution image fusion. Meanwhile, Ma et al. proposed GANMcC [33] where a multi-classification constrained adversarial network with main and auxiliary loss functions was designed to balance the gradient and intensity information. Although these GAN-based methods have achieved good performance, they still have limited ability in terms of highlighting thermal targets and unambiguous visible details. Ulteriorly, Yang \textit{et al.} [34] constructed a texture conditional generative adversarial network to capture texture map, and further proposed the squeeze-and-excitation module to highlight texture information. Li \textit{et al.} presented a multi-grained attentional network, namely MgAN-Fuse [35], which integrated attention modules into the encoder-decoder network to capture the context information in the generator. Meanwhile, they also introduced AttentionFGAN [36] where a multi-scale attention module was integrated into both generator and discriminator.

The above-mentioned methods mainly depend on the convolutional layer to accomplish local feature extraction and reconstruction, and emphasize on the elaborate design of network architecture, such as dense block [12, 31], residual block [26, 28] and multi-scale characteristic [25, 27, 29], etc. Furthermore, some of them introduce the attention mechanism into the convolutional neural network to improve feature representation ability [25-27]. In particular, Qu \textit{et al.} developed TransMEF [37] for multi-exposure image fusion, which integrated a CNN module and a transformer module to extract both local and global features. However, their method proposes the transformer as a supplement of CNN. Different from the existing methods, we introduce a pure transformer encoding backbone without the help of convolutional neural networks. With the stronger representation ability of the self-attention mechanism, our SwinFuse can bring a significant breakthrough for image fusion performance.

\section{Method}
In this section, we firstly introduce the network architecture, and then emphasize on the design of residual Swin Transformer block, fusion strategy and loss function.

\begin{figure*}[!t]
	\centering
	\includegraphics[width=1\textwidth]{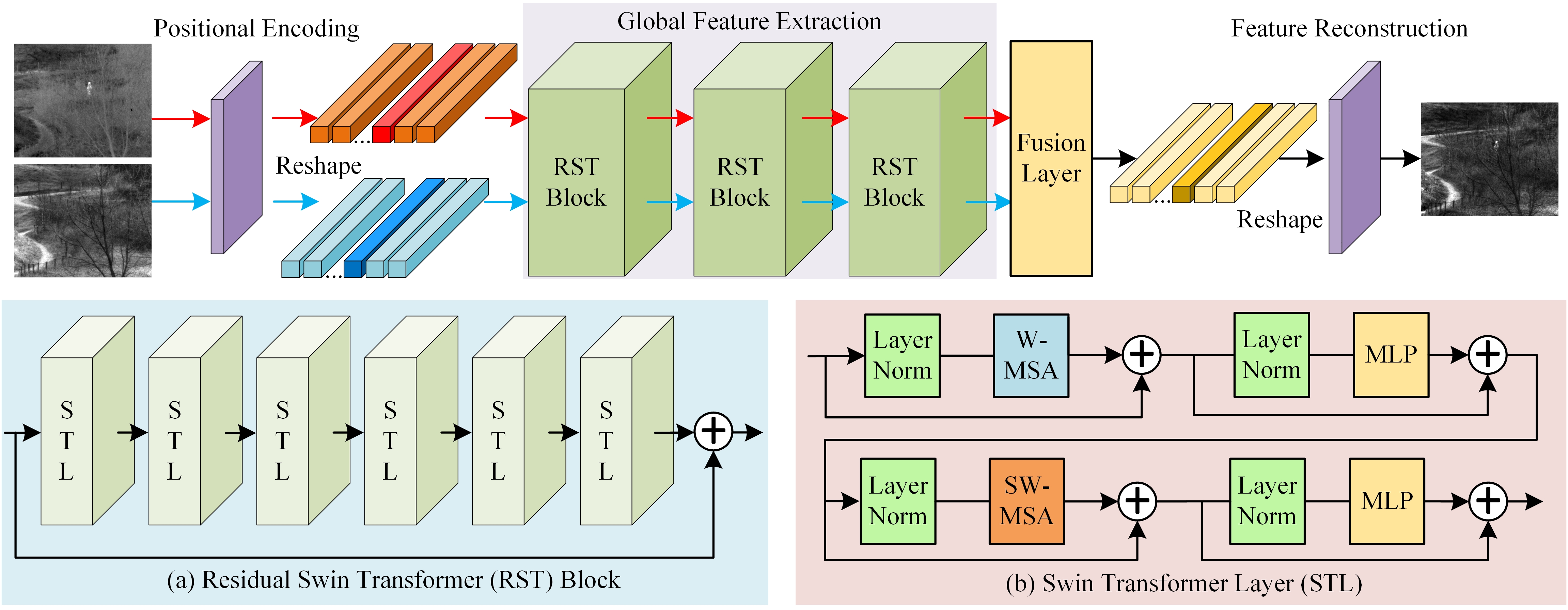}
	% where an .eps filename suffix will be assumed under latex, 
	% and a .pdf suffix will be assumed for pdflatex; or what has been declared
	% via \DeclareGraphicsExtensions.
	\caption{The network architecture of our SwinFuse, which consists of the global feature extraction, fusion layer and feature reconstruction. Notably, in the training phase, the fusion layer need to be removed.}
	\label{Fig2}
\end{figure*}

\subsection{Network Overview}

As illustrated as Fig.2, our SwinFuse consists of three main parts: global feature extraction, fusion layer and feature reconstruction. Given the testing infrared and visible images ${I^{l}} \in {R^{H \times W \times {C_{in}}}}\ $ (H, W and $ C_{in} $ respectively represent the height, width and input channel number, $l $=ir for infrared image, and $ l$=vis for visible image), we firstly use a convolutional layer with 1$ \times $1 kernel to implement positional encoding, and transform the channle from $ C_{in} $ to $ C $. The initial features $ {\Phi^l} \in {R^{M \times N \times C}} $ are defined by Eq.1.
\begin{equation}
\Phi ^l = H_{Pos}({I^l})
\end{equation}
where $ H_{Pos} $ denotes the positional encoding, $ C $ is the output channel number and set to 96. Notably, the convolution layer is an effective way for positional encoding, and transform an image space into a high-dimensional feature space. Subsequently, we reshape the initial features $ {\Phi^l}$ to sequence vectors $ {\Phi_{SV}^l}\in {R^{MN \times C}}$, and apply the residual Swin Transformer blocks (RSTBs) to extract the global features $ {\Phi_{GF}^l}\in {R^{MN \times C}}$, which are expressed by Eq.2. 
\begin{equation}
	\Phi _{GF}^l = H_{RSTB_m}({\Phi_{SV}^l})
\end{equation}
where $ H_{RSTB_m} $ represents the \textit{m}-th RSTB. With these operations, the global features of infrared and visible images are extracted. Then, we adopt a fusion layer based on $L_{1}$-norm from row and column vector dimensions to obtain the fused global features $ {\Phi_{F}}\in {R^{MN \times C}}$, which is formulated by Eq.3.
\begin{equation}
	\Phi _{F} = H_{Norm}({\Phi_{GF}^l})
\end{equation}
where $ H_{Norm} $ denotes the fusion operation. Finally, we again reshape the fused global features from $ R^{MN \times C} $ to $ R^{M \times N \times C} $, and use a convolutional layer to reconstruct a fusion image $ I_F $, which is defined by Eq.4.
\begin{equation}
	I_F  = H_{Conv}({\Phi _{F}})
\end{equation}
where $ H_{Conv} $ represents the feature reconstruction. This convolutional layer is with 1$ \times $1 kernel, a padding of 0 and along with a Tanh activation layer.

\begin{figure}[!t]
	\centering
	\includegraphics[width=0.45\textwidth]{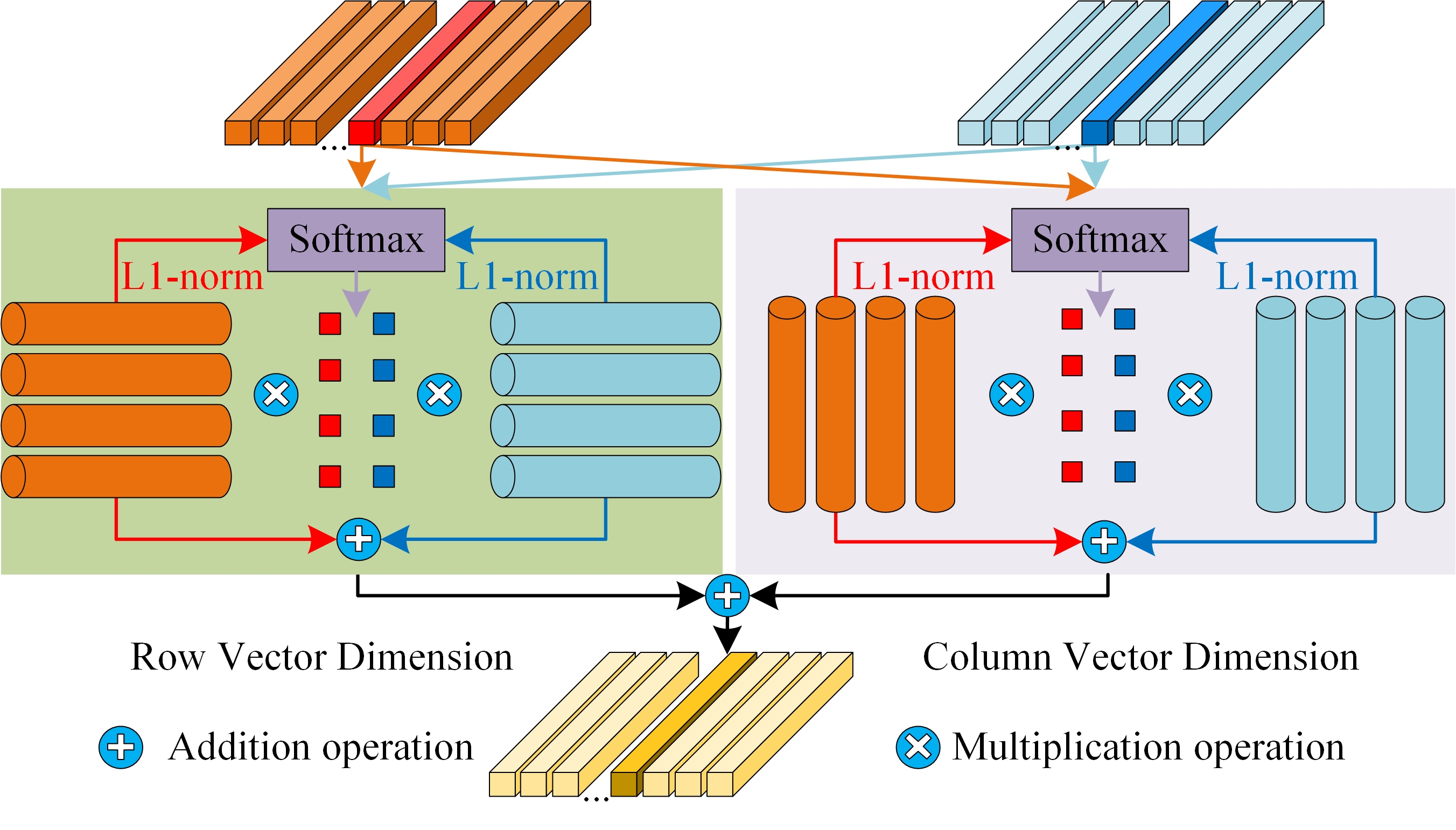}
	% where an .eps filename suffix will be assumed under latex, 
	% and a .pdf suffix will be assumed for pdflatex; or what has been declared
	% via \DeclareGraphicsExtensions.
	\caption{The fusion strategy of our SwinFuse. The left part performs row vetor normalization, while the right part performs column vetor normalization.}
	\label{Fig3}
\end{figure}

\subsection{Residual Swin Transformer Block}

Fig.2(a) describes the architecture of residual Swin Transformer block (RSTB), which includes a series of Swin Transformer layers (STLs) along with a residual connection. Given the input sequence vectors $ {\Phi_{m,0}^l}$, we apply \textit{n} Swin Transformer layers to extract the intermediate global features $ \Phi _{m,n-1}^l $, and the final output of RSTB is calculated by Eq.5.
\begin{equation}
	\Phi _{m,n}^l  = H_{STL_{m,n}}({\Phi _{m,n-1}^l})+\Phi_{m,0}^l
\end{equation}
where $ H_{STL_{m,n}} $ denotes the \textit{n}-th Swin Transformer layer. Similar to the CNN architecture, the multi-layer Swin Transformer can effectively model the global features, and residual connection can aggregate the different levels of features.

The Swin Transformer layer, which is shown in Fig.2(b), first utilizes the N$ \times $N sliding window to partition the input into the non-overlapping $ \frac{{HW}}{{{N^2}}} $ local windows, and computes their local attention. For the feature of local window $ \Phi_z $, the matrices Q, K and V are calculated by Eq.6.
\begin{equation}
 Q = {\Phi _z}{W_Q},   K = {\Phi _z}{W_K},   V = {\Phi _z}{W_V}
\end{equation}
where $ {W_Q} $, $ {W_K} $ and $ {W_V} \in R^{N^2 \times d} $ are the learnable parameters of three linear projection layers with sharing across different windows, and \textit{d} is the dimension of (Q, K). Meanwhile, the sequence matrices of self-attention mechanism are formulated by Eq.7.
\begin{equation}
	Attention(Q,K,V) = SoftMax ({{Q{K^T}} \mathord{\left/
			{\vphantom {{Q{K^T}} {\sqrt d }}} \right.
			\kern-\nulldelimiterspace} {\sqrt d }} + p)V
\end{equation}
where p is a learnable parameter for the positional decoding. Subsequently, the Swin Transformer layer computes again the standard multi-head self-attention (MSA) for the shifted windows. On the whole, it consists of a W-MSA and a SW-MSA, following by the multi-layer perceptron (MLP) with gaussian error linear units (GELU) nonlinearity in between them. A LayerNorm layer is applied before each of MSA and NLP, and a residual connection is employed for each module.

\subsection{Fusion Strategy}

In the fusion layer, as illustrated as Fig.3, we design a novel fusion strategy based on $L_{1}$-norm for the sequence matrices of infrared and visible images, and measure their activity level from row and column vector dimensions. For their respective global features, termed as $ {\Phi_{GF}^{ir}(i,j)}$ and $ {\Phi_{GF}^{vis}(i,j)}$, we firstly calcuate their row vector weights by $L_{1}$-norm, and adopt softmax to obtain their activity level, termed as $ \varphi _{row}^{ir}(i) $ and $ \varphi _{row}^{vis}(i)  $, which are expressed by Eq.8 and 9.
\begin{equation}
	\varphi _{row}^{ir}(i) = \frac{{\exp ({{\left\| {\Phi _{GF}^{ir}(i)} \right\|}_1})}}{{\exp ({{\left\| {\Phi _{GF}^{ir}(i)} \right\|}_1}) + \exp ({{\left\| {\Phi _{GF}^{vis}(i)} \right\|}_1})}}
\end{equation}
\begin{equation}
	\varphi _{row}^{vis}(i) = \frac{{\exp ({{\left\| {\Phi _{GF}^{vis}(i)} \right\|}_1})}}{{\exp ({{\left\| {\Phi _{GF}^{ir}(i)} \right\|}_1}) + \exp ({{\left\| {\Phi _{GF}^{vis}(i)} \right\|}_1})}}
\end{equation}
where $ \left\|  \cdot  \right\|_1 $ denotes the $L_{1}$-norm calculatation. Then, we directly multiply their activity level with the corresponding global features to obtain the fused global features from row vector dimension, termed as $ \Phi _{F}^{row}(i,j) $, which is formulated by Eq.10.
\begin{equation}
    \Phi _{F}^{row}(i,j) = \Phi _{GF}^{ir}(i,j) \times \varphi _{row}^{ir}(i) + \Phi _{GF}^{vis}(i,j) \times \varphi _{row}^{vis}(i)
\end{equation}

Subsequently, similar to the above operations, we measure their activity level from column vector dimension, termed as $ \varphi _{col}^{ir}(j) $ and $ \varphi _{col}^{vis}(j)  $, which are expressed by Eq.11 and 12.
\begin{equation}
	\varphi _{col}^{ir}(j) = \frac{{\exp ({{\left\| {\Phi _{GF}^{ir}(j)} \right\|}_1})}}{{\exp ({{\left\| {\Phi _{GF}^{ir}(j)} \right\|}_1}) + \exp ({{\left\| {\Phi _{GF}^{vis}(j)} \right\|}_1})}}
\end{equation}
\begin{equation}
	\varphi _{col}^{vis}(j) = \frac{{\exp ({{\left\| {\Phi _{GF}^{vis}(j)} \right\|}_1})}}{{\exp ({{\left\| {\Phi _{GF}^{ir}(j)} \right\|}_1}) + \exp ({{\left\| {\Phi _{GF}^{vis}(j)} \right\|}_1})}}
\end{equation}

And then, we can obtain the fused global features with column vector dimension, termed as $ \Phi _F^{col}(i,j) $, which is formulated by Eq.13.
\begin{equation}
	\Phi _F^{col}(i,j) = \Phi _{GF}^{ir}(i,j) \times \varphi _{col}^{ir}(j) + \Phi _{GF}^{vis}(i,j) \times \varphi _{col}^{vis}(j)
\end{equation}

Finally, we adopt element-wise addition operation for their fused global features in the row and column vector dimensions, and obtain the final fused global features, which is calcuated by Eq.14.
\begin{equation}
	\Phi_F(i,j)= \Phi _F^{row}(i,j) + \Phi _F^{col}(i,j) 
\end{equation}

The obtained final fused global features are used to reconstruct the fusion image by a convolutional layer. It is worth noting that the fusion layer is only retained during the testing phase while removed during the training phase.

\subsection{Loss Function}

In the training phase, we propose the structure similarity (SSIM) and $L_{1}$ as the loss functions to supervise the network training. The SSIM is defined as independent of image brightness and contrast, and reflects the attributes of structure information, such as scene details and structural texture, but it is prone to color deviation and brightness variation. Therefore, we again adopt the $L_{1}$ loss function to make up for its shortcomings. Therefore, the SSIM and $L_{1}$ loss functions are respectively defined as Eq.15 and 16.
\begin{equation}
	{L_{ssim}} = 1 - SSIM({I_{out}} - {I_{in}})
\end{equation}
\begin{equation}
	{L_{l1}} = \frac{1}{{HW}}\sum {\left| {{I_{out}} - {I_{in}}} \right|}
\end{equation}
where $ SSIM(\cdot) $ denotes the SSIM operation, $ I_{in} $ and $ I_{out} $ respectively represent the input and output images. Moreover, the total loss function is defined by Eq.17.
\begin{equation}
	{L_{total}} = 	{L_{l1}}+\lambda {L_{ssim}}
\end{equation}
where $ \lambda $ is a hyper-parameter, and used to adjust the difference of order magnitude between $ {L_{l1}} $ and $  {L_{ssim}} $. In the section IV, we will discuss its impact on the fusion performance.

\section{Experiments and Analyses}
In this section, we firstly introduce the experimental setup, and then focus on the discussion and analysis of the relevant experiments.

\subsection{Experimental Setup}
During the training phase, we propose the MS-COCO [38] dataset, which consists of more than 80000 natural images with different categories, to train our SwinFuse network. To accommodate the network training, all images are transformed into the size of 224×224 and grayscale range [-1, 1]. Moreover, the numbers of RSTB and STL are set to 3 and 6. The patch size and sliding window size are set to 1$ \times $1 and 7$ \times $7. The head numbers of three RSTLs are set to 1, 2 and 4, respectively. In addition, we use the Adam as a optimizer, and set the learning rate, batchsize and epoch to 1$\times$ $ {10^{ - 5}} $, 4 and 50, respectively. The training platform is with Intel I9-10850K CPU, 64GB RAM, NVIDIA GeForce GTX 3090 GPU.

During the testing phase, we adopt three datasets, namely TNO [15], Roadscene [39] and OTCBVS [40], to demonstrate the effectiveness of our SwinFuse, and successively select 20, 40 and 31 images from the corresponding datasets. Moreover, we transform the grayscale range of source images to -1 and 1, and ultilize the sliding window 224$ \times $224 to partition them into several patches, where the value of invalid region is filled with 0. After the combination of each patch pair, we perform the reverse operation according to the previous partition order to obtain the final fusion image.

Meanwhile, we choose nine representative methods, namely, MDLatLRR [5], IFCNN [24], DenseFuse [12], RFN-Nest [28], FusionGAN [13], GANMcC [33], PMGI [30], SEDRFuse [25] and Res2Fusion [26] to compared with our SwinFuse. Meanwhile, eight evaluation indexs, namely average gradient (AG), spatial frequency (SF) [41], standard deviation (SD) [42], multi-scale structure similarity (MS\_SSIM) [43], feature mutual information by wavelet (FMI\_w) [44], mutual information (MI) [45], the sum of the correlation differences (SCD) [46] and visual information fidelity for fusion (VIFF) [47], are selected for the fair and comprehensive comparisons.

\begin{figure*}[!t]
	\centering
	\includegraphics[width=1\textwidth]{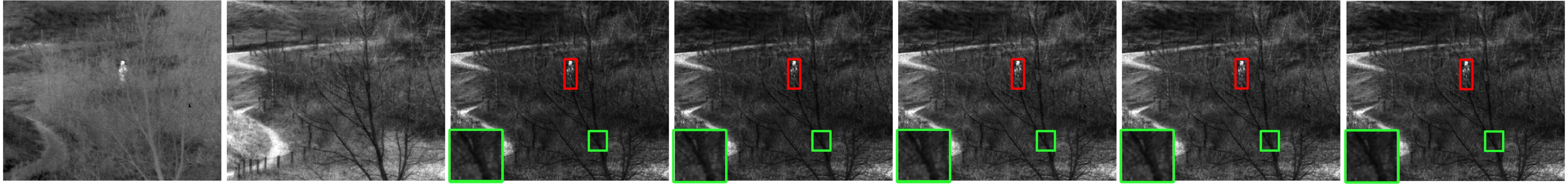}
	% where an .eps filename suffix will be assumed under latex, 
	% and a .pdf suffix will be assumed for pdflatex; or what has been declared
	% via \DeclareGraphicsExtensions.
	\caption{The subjective ablation comparisons of \textit{Sandpath} selected from the TNO dataset. These images successively are source images and the results obtained by the proposed SwinFuse with different parameters $ \lambda $, respectively.}
	\label{Fig4}
\end{figure*}

\begin{table*}[!t]
	\renewcommand\arraystretch{1.3}
	%% increase table row spacing, adjust to taste
	%\renewcommand{\arraystretch}{1.3}
	% if using array.sty, it might be a good idea to tweak the value of
	% \extrarowheight as needed to properly center the text within the cells
	\caption{The objective ablation comparisons of the TNO dataset for the different parameters $ \lambda $.}
	\label{table1}
	\centering
	%% Some packages, such as MDW tools, offer better commands for making tables
	%% than the plain LaTeX2e tabular which is used here.
	\begin{tabular}{ l c c c c c c c c}
		\hline
		Parameters & AG & SF & SD & MI &  MS\_SSIM  & FMI\_w & SCD & VIFF \\
		\hline
		$\lambda=$1e0 & 6.03910 & 12.74940 & 46.70967 & \color{blue}{2.31879} & 0.91318 & 0.42281 & 1.80277 & 0.72934 
		\\
		$\lambda=$1e1 & 5.99930 & 12.58878 & 46.29191 & 2.26808 & 0.91356 & 0.42102 & 1.80701 & 0.72750 
		\\
		$\lambda=$1e2 & \color{red}{6.19931} & \color{blue}{12.75545} & \color{blue}{46.88304} & 2.29560 & \color{blue}{0.92020} & \color{blue}{0.42521} & 1.81779 & \color{red}{0.76083} 
		\\
		$\lambda=$1e3 & \color{blue}{6.19038} & \color{red}{12.79886} & \color{red}{46.90388} & \color{red}{2.32346} & \color{red}{0.92066} & \color{red}{0.42618} & \color{red}{1.84127} & \color{blue}{0.76068} 
		\\
		$\lambda=$1e4 & 6.12769 & 12.72406 & 46.54066 & 2.24862 & 0.91846 & 0.42361 & \color{blue}{1.83106} & 0.74579 
		\\
		\hline 
	\end{tabular}
\end{table*}

\begin{figure*}[!t]
	\centering
	\includegraphics[width=1\textwidth]{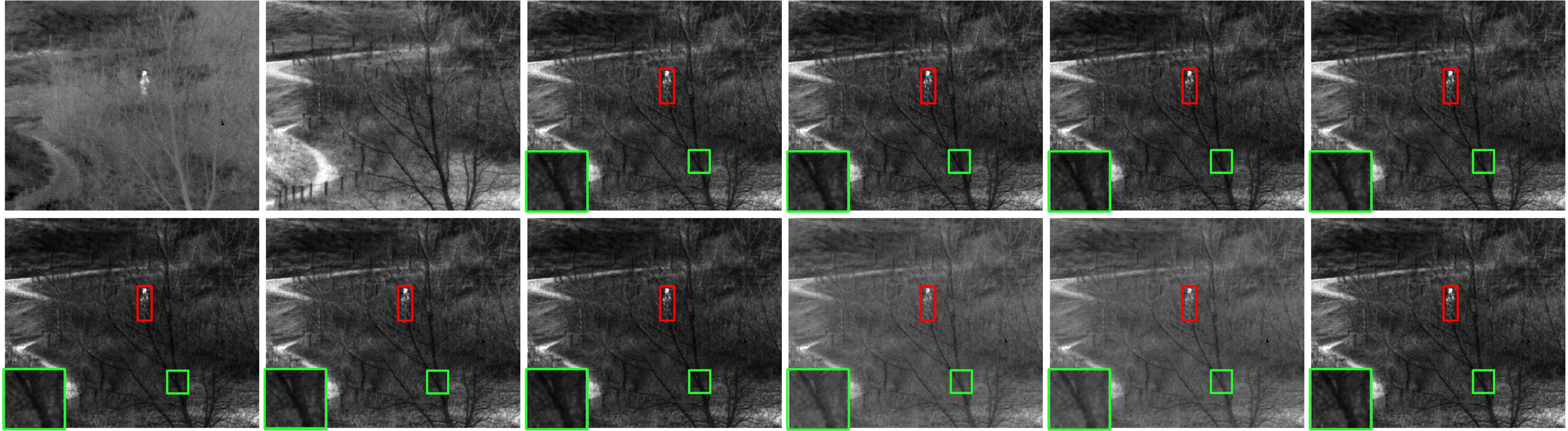}
	% where an .eps filename suffix will be assumed under latex, 
	% and a .pdf suffix will be assumed for pdflatex; or what has been declared
	% via \DeclareGraphicsExtensions.
	\caption{The subjective ablation comparisons of \textit{Sandpath} for the RSTL framework. These images successively are source images, the fused results of RSTBs with \textit{m}=2, 4, 5, STLs with \textit{n}=5, 7, 8, without residual connection, Only\_row, Only\_col and the proposed SwinFuse, respectively.}
	\label{Fig5}
\end{figure*}

\begin{table*}[!t]
	\renewcommand\arraystretch{1.3}
	\newcommand{\tabincell}[2]{\begin{tabular}{@{}#1@{}}#2\end{tabular}} 
	%% increase table row spacing, adjust to taste
	%\renewcommand{\arraystretch}{1.3}
	% if using array.sty, it might be a good idea to tweak the value of
	% \extrarowheight as needed to properly center the text within the cells
	\caption{The objective ablation comparisons of the TNO dataset for the RSTL framework.}
	\label{table2}
	\centering
	%% Some packages, such as MDW tools, offer better commands for making tables
	%% than the plain LaTeX2e tabular which is used here.
	\begin{tabular}{ c c c c c c c c c c}
		\hline
		Models & Parameters & AG & SF & SD & MI &  MS\_SSIM  & FMI\_w & SCD & VIFF \\
		\hline
		\multirow{4}{*}{\tabincell{c}{RSTB \\Number}}& 2 & 6.12630 & 12.78277 & 46.64155 
		& 2.28553 & \color{blue}{0.92034} & \color{blue}{0.42579} & \color{blue}{1.84013} & \color{blue}{0.76044}   \\
		&3 & \color{red}{6.19038} & \color{blue}{12.79886} & \color{red}{46.90388} & \color{blue}{2.32346} & \color{red}{0.92066} & \color{red}{0.42618} & \color{red}{1.84127} & \color{red}{0.76068} 	\\
		&4 & 6.12761 & \color{red}{12.83733} & \color{blue}{46.87729} & 2.30659 & 0.91471 & 0.42474 & 1.81931 & 0.73768   	\\
		&5 & \color{blue}{6.15424} & 12.78313 & 46.83528 & \color{red}{2.33390} & 0.91943 & 0.42503 & 1.83268 & 0.75333   \\
		\hline
		\multirow{4}{*}{\tabincell{c}{STL \\Number}} &5 & \color{blue}{6.18744} & 12.78088 & \color{blue}{46.65097} & \color{blue}{2.29336} 
		& 0.91996 & 0.42522 & \color{red}{1.84265} & 0.75654    	\\
		&6 & \color{red}{6.19038} & \color{blue}{12.79886} & \color{red}{46.90388}  & \color{red}{2.32346} & \color{blue}{0.92066} & \color{red}{0.42618} & \color{blue}{1.84127} & \color{red}{0.76068} 	\\
		&7 & 6.07115 & \color{red}{12.80509} & 45.74293  & 2.21655 & 0.90937 & 0.42057 & 1.80394 & 0.72002  	\\
		&8 & 6.09115 & 12.78692 & 46.47507  & 2.23655 & \color{red}{0.92072} & \color{blue}{0.42582} & 1.82394 & \color{blue}{0.75977}   \\
		\hline 
		\multirow{2}{*}{\tabincell{c}{Residual\\Connection}} &No & 6.06763 & 12.77942 & 46.32200 
		& \color{red}{2.32840}  & 0.91464 & 0.42297 & 1.81308 & 0.73333  	      	\\
		&Yes & \color{red}{6.19038} & \color{red}{12.79886} & \color{red}{46.90388} & 2.32346 & \color{red}{0.92066} & \color{red}{0.42618}  & \color{red}{1.84127} & \color{red}{0.76068}   \\
		\hline 
		\multirow{3}{*}{\tabincell{c}{Fusion\\Layer}} &Only\_row & \color{blue}{4.88061} & \color{blue}{10.26035} & \color{blue}{35.62175} 
		& \color{blue}{2.32627} & 0.90146 & \color{blue}{0.42048} & 1.74474 & 0.49377 
		\\
		&Only\_col & 4.87315 & 10.21141 & 34.90279 & \color{red}{2.38689} & \color{blue}{0.90309} & 0.42013 & \color{blue}{1.74862} & \color{blue}{0.49817} 
		\\
		&Ours & \color{red}{6.19038} & \color{red}{12.79886} & \color{red}{46.90388} & 2.32346 & \color{red}{0.92066} & \color{red}{0.42618}  & \color{red}{1.84127} & \color{red}{0.76068}   \\
		\hline 
	\end{tabular}
\end{table*}

\subsection{Ablation Study}

\subsubsection{The impact of parameter $ \lambda $} In the design of loss function, we apply a hyper-parameter $ \lambda $ to balance the difference of order magnitude between $ {L_{l1}} $ and $  {L_{ssim}} $. Therefore, in this ablation study, we set $ \lambda $ to 1(1e0), 10(1e1), 100(1e2), 1000(1e3) and 10000(1e4) to verify the impact of different parameters on the fusion performance. The above-mentioned TNO dataset and eight evaluation indexes are selected for the experimental verification, and the corresponding optimal and suboptimal average values of evaluation indexes are labeled by red and blue.

\begin{figure*}[!t]
	\centering
	\includegraphics[width=1\textwidth]{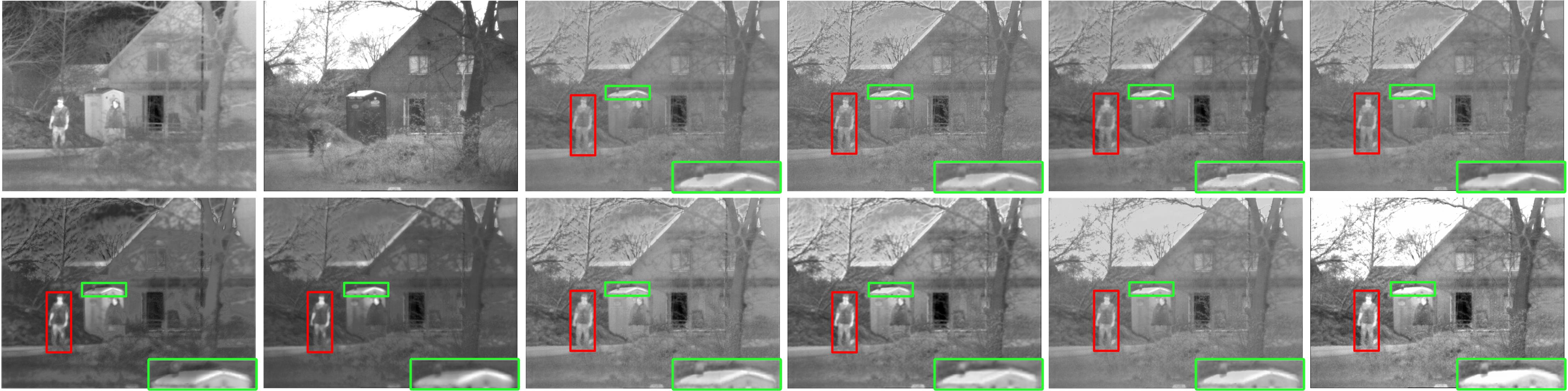}
	% where an .eps filename suffix will be assumed under latex, 
	% and a .pdf suffix will be assumed for pdflatex; or what has been declared
	% via \DeclareGraphicsExtensions.
	\caption{The subjective comparisons of \textit{2\_{men}\_{in}\_{front}\_{of}\_{house}} selected from TNO dataset. These images are source images, the results of MDLatLRR [5], IFCNN [24], DenseFuse [12], RFN-Nest [28], FusionGAN [13], GANMcC [33], PMGI [30], SEDRFuse [25] and Res2Fusion [26] and our SwinFuse, respectively.}
	\label{Fig6}
\end{figure*}

\begin{figure*}[!t]
	\centering
	\includegraphics[width=1\textwidth]{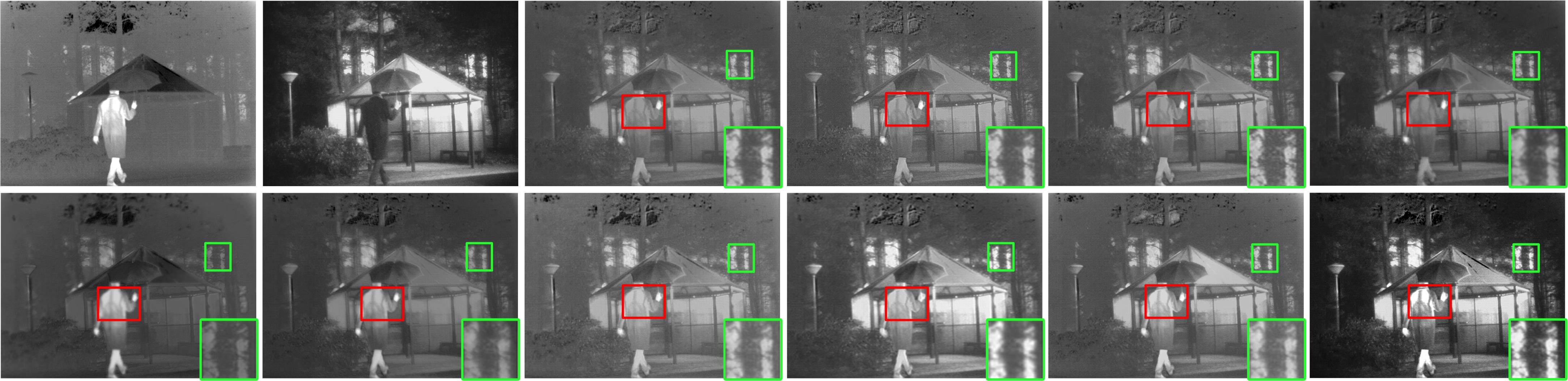}
	% where an .eps filename suffix will be assumed under latex, 
	% and a .pdf suffix will be assumed for pdflatex; or what has been declared
	% via \DeclareGraphicsExtensions.
	\caption{The subjective comparisons of \textit{Kaptein\_{1654}} selected from TNO dataset. These images are source images, the results of MDLatLRR [5], IFCNN [24], DenseFuse [12], RFN-Nest [28], FusionGAN [13], GANMcC [33], PMGI [30], SEDRFuse [25] and Res2Fusion [26] and our SwinFuse, respectively.}
	\label{Fig7}
\end{figure*}

Fig.4 shows the subjective ablation comparisons of \textit{Sandpath} for our SwinFuse with different parameters. From the visual effect, the disparity of these results is very weak, especially for the labeled typical target and local details. Meanwhile, their objective ablation results are presented in Table I. Our SwinFuse with $ \lambda $=1e3 achieves the optimal values of SF, SD, MI, MS\_SSIM, FMI\_W and SCD, and the suboptimal values of AG and VIFF, which follow behind that of $ \lambda $=1e2. This ablation study demonstrates that our SwinFuse with $ \lambda $=1e3 obtains the best performance, and propose it for the subsequent experimental verification.

\subsubsection{The impact of RSTL framework}
In the proposed network, our SwinFuse includes \textit{m} residual Swin Transformer blocks (RSTBs) and \textit{n} Swin Transformer layers (STLs). In this ablation study, we verify the numbers of RSTL and STL, as well as the residual connection, on the impact of the fusion performance. we set \textit{m} to 2, 3, 4, 5, and \textit{n} to 5, 6, 7, 8. Moreover, in the fusion layer, we also verify the impact of fusion strategy with only row vector dimension (termed as Only\_row) and only column vector dimension (termed as Only\_col). We select the TNO dataset for this ablation study.

Fig.5 gives the subjective comparisons of \textit{Sandpath}. These images successively are source images, the fused results of RSTBs with \textit{m}=2, 4, 5, STLs with \textit{n}=5, 7, 8, without residual connection, Only\_row, Only\_col and the proposed SwinFuse, respectively. We can find that the visual disparities of different RSTBs, STLs and without residual connection are inconspicuous. Nevertheless, the visual effect of Only\_row and Only\_col are relatively poor and have low contrast. However, the proposed SwinFuse with \textit{m}=3 and \textit{n}=6 retains conspicuous infrared targets and unambiguous visible details. 

In addition, Table II presents the objective ablation comparisons. In the design of RSTB and STL, when the number of RSTB is 3, our SwinFuse achieves the optimal values of AG, SD, MS\_SSIM, FMI\_w, SCD and VIFF, the suboptimal values of SF and MI. Moreover, when the number of STL is 6, our SwinFuse achieves the optimal values of AG, SD, MI, FMI\_w and VIFF, the suboptimal values of SF, MS\_SSIM and SCD. These results indicate that our SwinFuse has better fusion performance in the case of m=3 and n=6. Meanwhile, compared with without residual connection, Only\_row and Only\_col, the proposed SwinFuse achieve all the best indexes except for MI. The ablation studies indicate that the proposed network architecture is reasonable and effective.

\begin{figure*}[!t]
	\centering
	\includegraphics[width=1\textwidth]{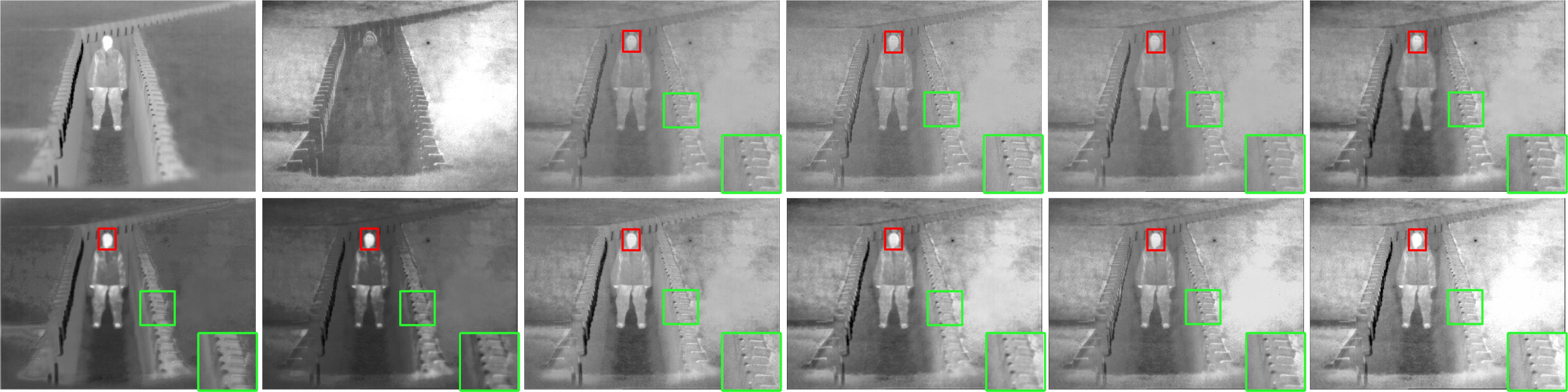}
	% where an .eps filename suffix will be assumed under latex, 
	% and a .pdf suffix will be assumed for pdflatex; or what has been declared
	% via \DeclareGraphicsExtensions.
	\caption{The subjective comparisons of \textit{soldier\_{in}\_{trench}\_1} selected from TNO dataset. These images are source images, the results of MDLatLRR [5], IFCNN [24], DenseFuse [12], RFN-Nest [28], FusionGAN [13], GANMcC [33], PMGI [30], SEDRFuse [25] and Res2Fusion [26] and our SwinFuse, respectively.}
	\label{Fig8}
\end{figure*}

\begin{figure*}[!t]
	\centering
	\includegraphics[width=1\textwidth]{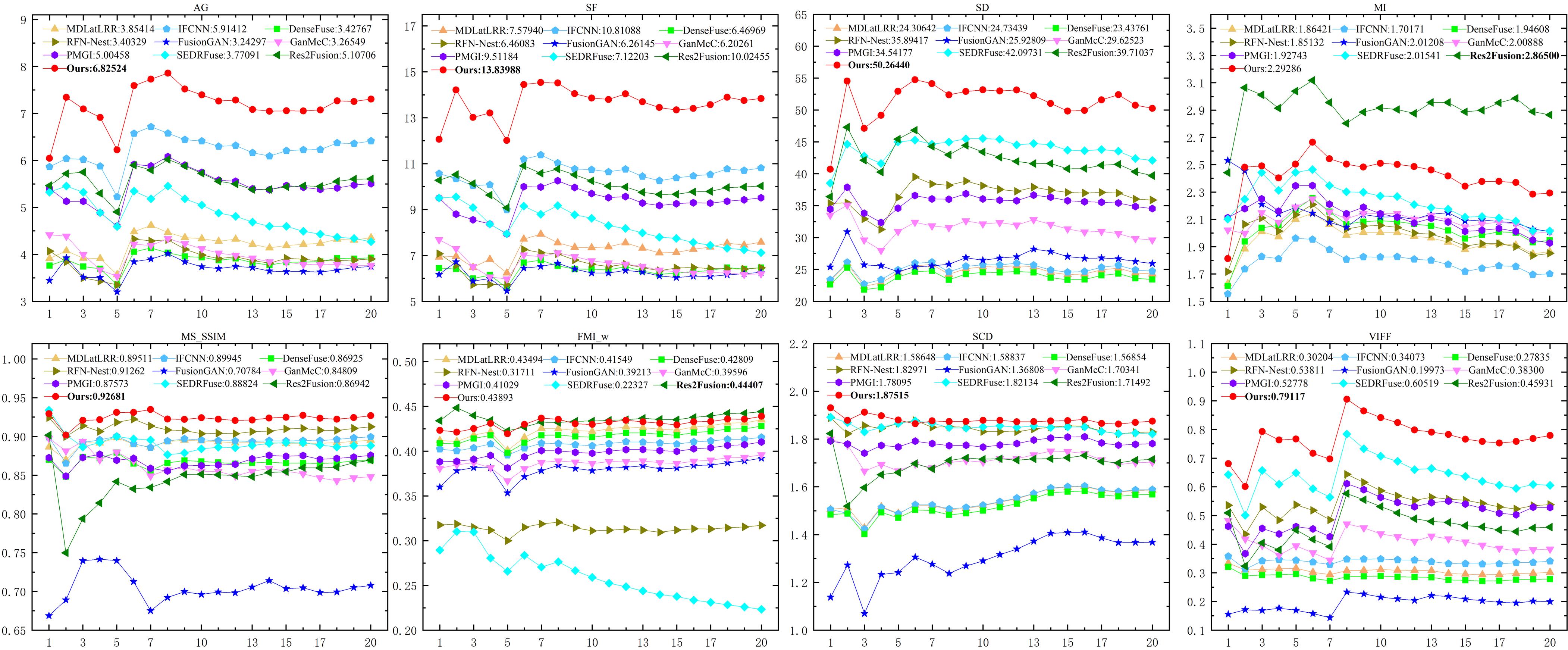}
	% where an .eps filename suffix will be assumed under latex, 
	% and a .pdf suffix will be assumed for pdflatex; or what has been declared
	% via \DeclareGraphicsExtensions.
	\caption{The objective comparisons of eight evaluation indexes with other nine methods for the TNO dataset. The red dotted line represents our SwinFuse.}
	\label{Fig9}
\end{figure*}

\subsection{Experiments on TNO Dataset}
To demonstrate the effectiveness of our SwinFuse, we carry out experiments on the TNO dataset. Three typical examples, such as \textit{{2\_{men}\_{in}\_{front}\_{of}\_{house}}},  \textit{Kaptein\_{1654}} and \textit{soldier\_{in}\_{trench}\_1}, are shown in Fig.6-8. The traditional MDLatLRR has a limited feature extraction ability by using the latent low-rank representation, and the obtained results have some serious loss of details and brightness. The IFCNN and DenseFuse propose a simple network and average addition fusion strategy, the obtained results prefer to preserve more visible details while losing the brightness of infrared targets. Moreover, the RFN-Nest adopts a multi-scale deep framework with two-stage training, its results still cannot retain the typical infrared targets. The FusionGAN and GANMcC introduce the adversarial learning mechanism, and the obtained results retain sharpening infrared targets, while the visible details are severely fuzzy and lacking. The PMGI design gradient and intensity paths for information extraction, and the obtained fusion performance is improved to a certain degree. In addition, the SEDRFuse and Res2Fusion accomplish reversely superior visual effect, the main reason is that these two methods introduce attention-based fusion strategies, and improve feature representation capability to some extent. However, compared with other methods, our SwinFuse obtains the best visual perception in maintaining visible details and infrared targets.

To better display the visual effect, we mark some representative targets and details with red and green boxes, and enlarge the marked local details. In the results of Fig.6-8, for the pedestrian targets and typical details, such as the roof of the toolhouse, trees and the edges of the trench, the MDLatLRR, IFCNN, DenseFuse and RFN-Nest can preserve these typical details from visible images, while damage the brightness of these pedestrian targets. Inversely, the FusionGAN and GANMcC can maintain these high-brightness pedestrians from infrared images, but generate some sharpened effect with blurred edges. More seriously, the important details of visible images are unclear and even missing. By contrast, the PMGI, SEDRFuse and Res2Fusion obtain better results, while their retention ability is still limited. However, our SwinFuse almost maintains the complete infrared targets with high brightness and unambiguous visible details. On the whole, our SwinFuse generates a better visual effect and achieves a higher contrast, conforming to human visual observation and other machine vision tasks.

Subsequently, we continue to verify the proposed SwinFuse through the objective index evaluation. Fig.9 presents the objective comparable results of eight evaluation indexes with other nine methods for the TNO dataset. Notably, the horizontal coordinate represents the number of testing images, while the vertical coordinate denotes the average values of evaluation index for the corresponding images. The results of our SwinFuse are represented by the red dotted line. From these results, our SwinFuse wins the first level for AG, SF, SD, MS\_SSIM, SCD and VIFF, the second level for MI and FMI\_w, which are inferior to Res2Fusion. The objective comparisons indicate that our SwinFuse generates more superior fusion performance than state-of-the-art traditional and deep learning based methods, which can draw the same conclusion as the above subjective analysis.

\begin{figure*}[!t]
	\centering
	\includegraphics[width=1\textwidth]{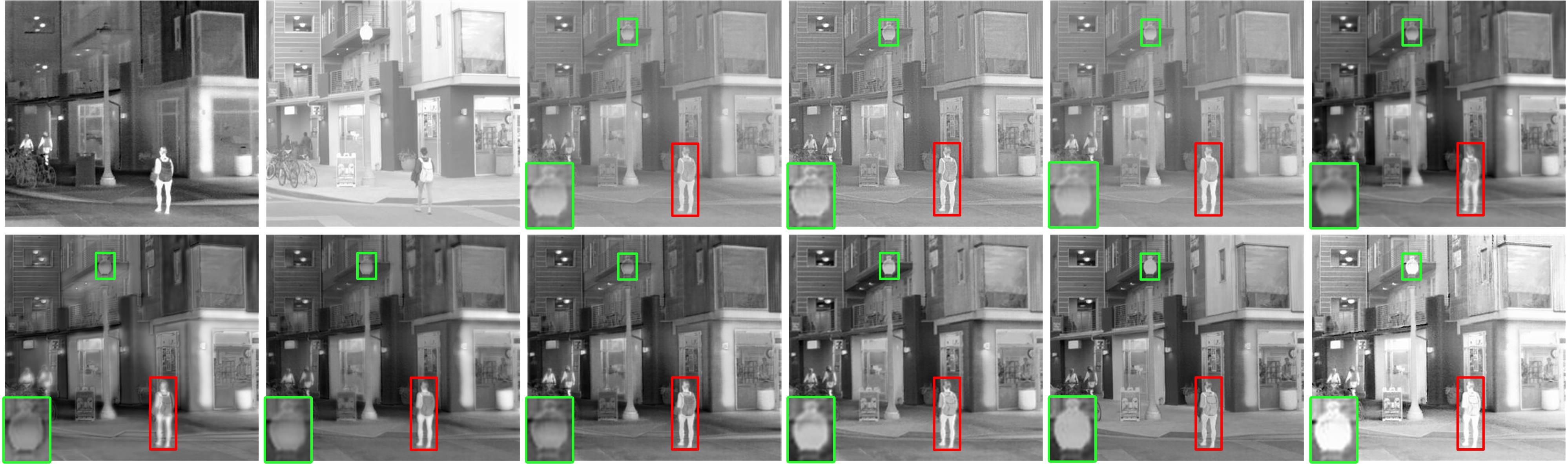}
	% where an .eps filename suffix will be assumed under latex, 
	% and a .pdf suffix will be assumed for pdflatex; or what has been declared
	% via \DeclareGraphicsExtensions.
	\caption{The subjective comparisons of \textit{FLIR\_08835} selected from Roadscene dataset. These images are source images, the results of MDLatLRR [5], IFCNN [24], DenseFuse [12], RFN-Nest [28], FusionGAN [13], GANMcC [33], PMGI [30], SEDRFuse [25] and Res2Fusion [26] and our SwinFuse, respectively.}
	\label{Fig10}
\end{figure*}

\begin{figure*}[!t]
	\centering
	\includegraphics[width=1\textwidth]{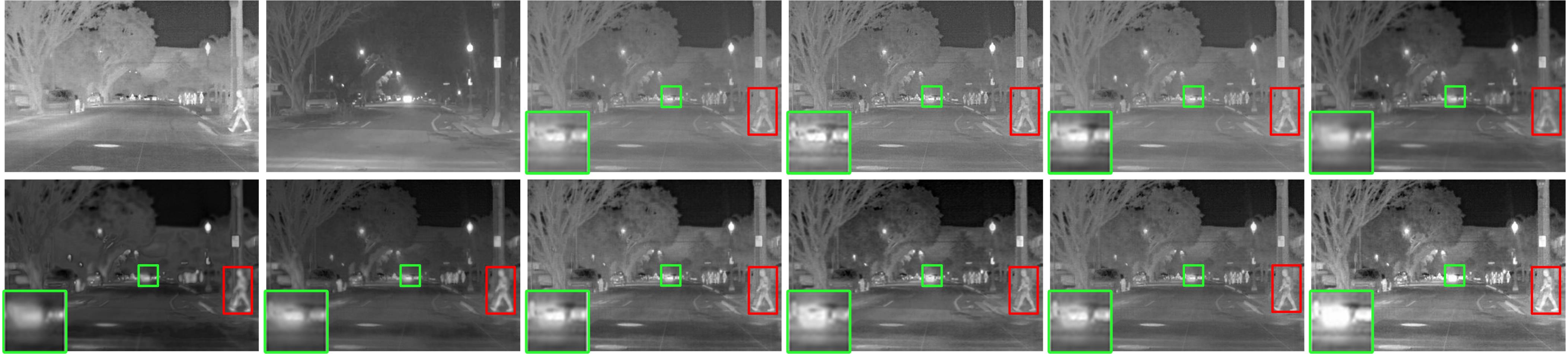}
	% where an .eps filename suffix will be assumed under latex, 
	% and a .pdf suffix will be assumed for pdflatex; or what has been declared
	% via \DeclareGraphicsExtensions.
	\caption{The subjective comparisons of \textit{FLIR\_08094} selected from Roadscene dataset. These images are source images, the results of MDLatLRR [5], IFCNN [24], DenseFuse [12], RFN-Nest [28], FusionGAN [13], GANMcC [33], PMGI [30], SEDRFuse [25] and Res2Fusion [26] and our SwinFuse, respectively.}
	\label{Fig11}
\end{figure*}

\begin{figure*}[!t]
	\centering
	\includegraphics[width=1\textwidth]{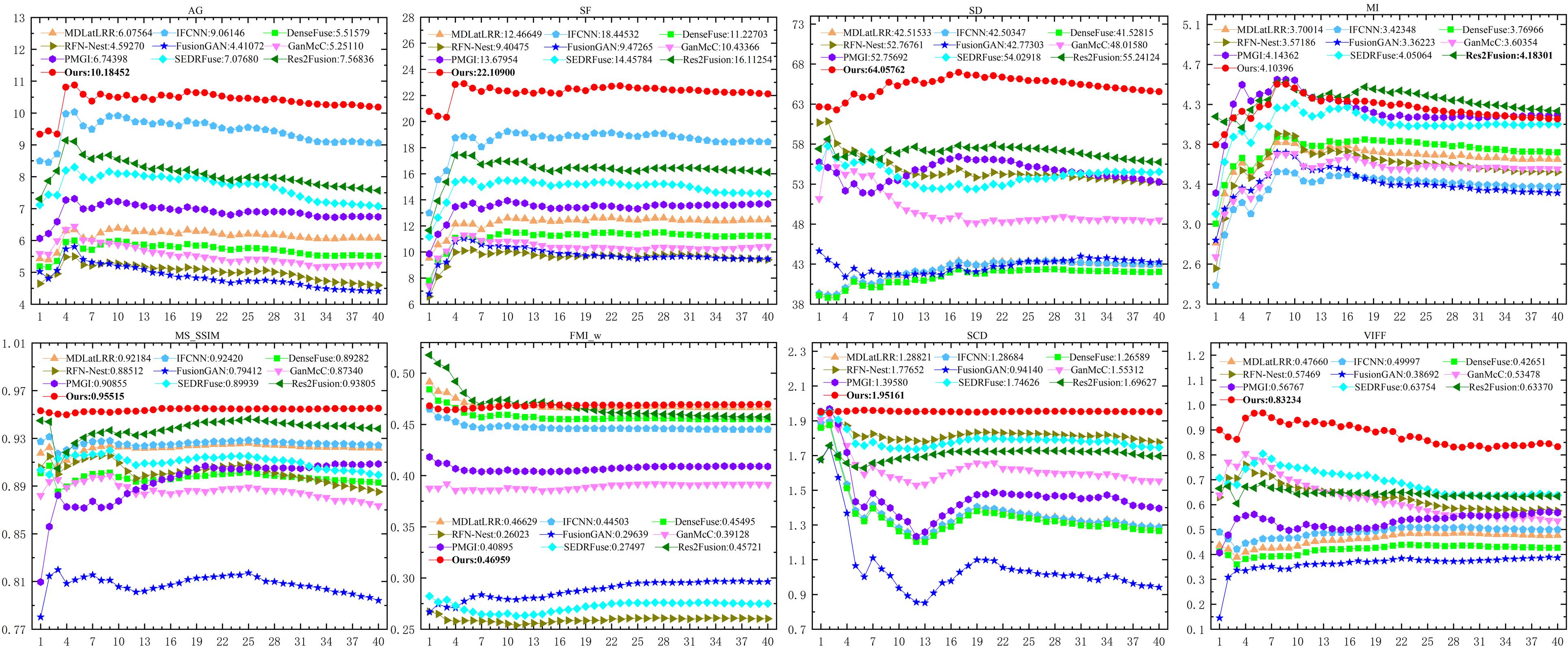}
	% where an .eps filename suffix will be assumed under latex, 
	% and a .pdf suffix will be assumed for pdflatex; or what has been declared
	% via \DeclareGraphicsExtensions.
	\caption{The objective comparisons of eight evaluation indexes with other nine methods for the Roadscene dataset. The red dotted line represents our SwinFuse.}
	\label{Fig12}
\end{figure*}

\begin{figure*}[!t]
	\centering
	\includegraphics[width=1\textwidth]{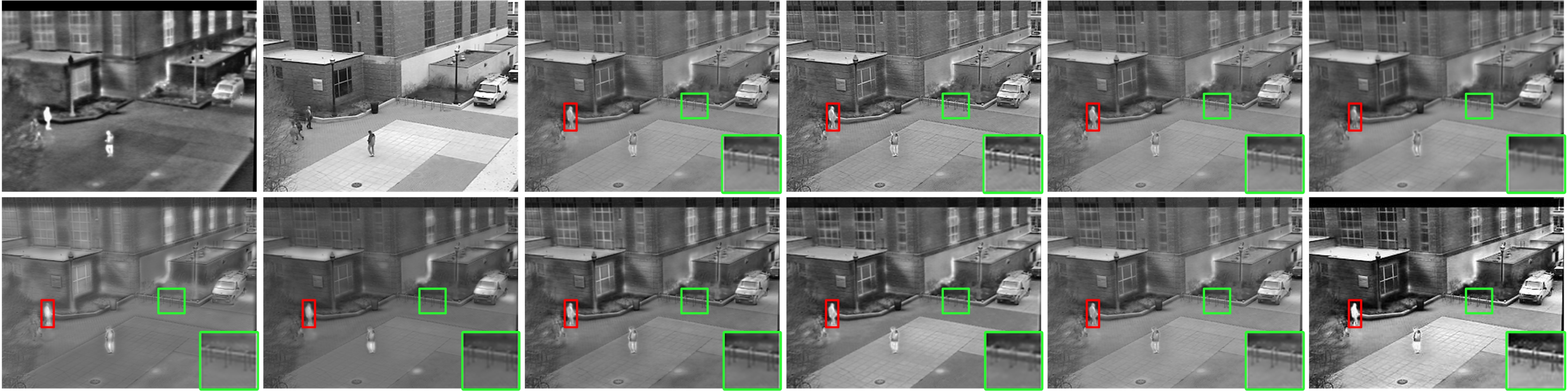}
	% where an .eps filename suffix will be assumed under latex, 
	% and a .pdf suffix will be assumed for pdflatex; or what has been declared
	% via \DeclareGraphicsExtensions.
	\caption{The subjective comparisons of \textit{video\_1036} selected from OTCBVS dataset. These images are source images, the results of MDLatLRR [5], IFCNN [24], DenseFuse [12], RFN-Nest [28], FusionGAN [13], GANMcC [33], PMGI [30], SEDRFuse [25] and Res2Fusion [26] and our SwinFuse, respectively.}
	\label{Fig13}
\end{figure*}

\begin{figure*}[!t]
	\centering
	\includegraphics[width=1\textwidth]{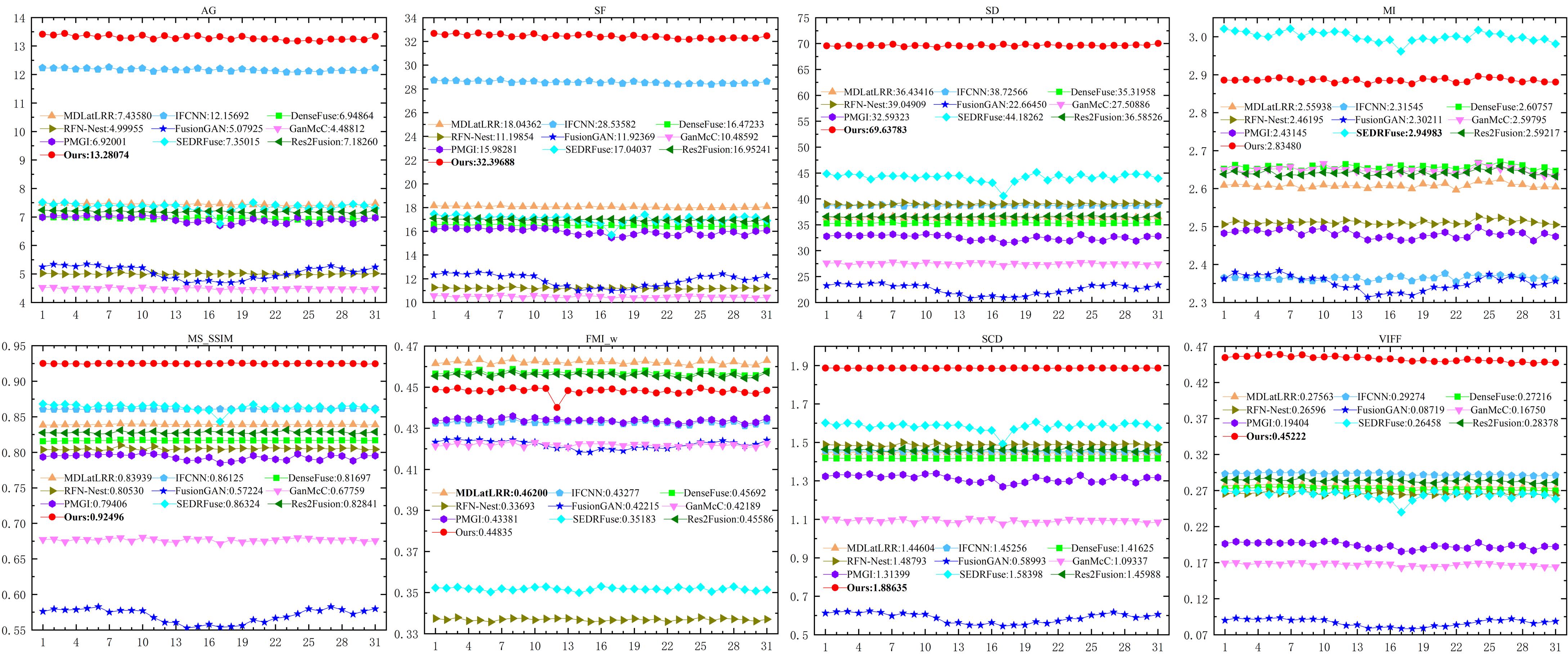}
	% where an .eps filename suffix will be assumed under latex, 
	% and a .pdf suffix will be assumed for pdflatex; or what has been declared
	% via \DeclareGraphicsExtensions.
	\caption{The objective comparisons of eight evalution indexes with other nine methods for the OTCBVS dataset. The red dotted line represents our SwinFuse.}
	\label{Fig14}
\end{figure*}

\subsection{Experiments on Roadscene Dataset}
Next, we further testify the proposed SwinFuse on the Roadscene dataset, in which 40 image pairs are selected for the testing. Fig.10 and Fig.11 describe the subjective comparisons of two examples, \textit{i.e,} \textit{FLIR\_08835} and \textit{FLIR\_08094}. For the typical pedestrian targets, our SwinFuse achieves higher brightness in the fusion image compared with other nine methods. Similarly, for the typical details, \textit{i.e,} street lamp and car, the results of our SwinFuse are complete and clear. Meanwhile, Fig.12 shows the objective experimental results. Our SwinFuse wins the first bank for AG, SF, SD, MS\_SSIM, FMI\_w, SCD and VIFF, and the suboptimal value of MI, which is in arrears of that of Res2Fusion. The subjective and objective experiments demonstrate our SwinFuse is more superior and transcends other methods.

In addition, from these objective evaluation indexes, the largest SCD and FMI\_w indicate that our fused results maintain more similar feature retention from source images. The largest SF and MS\_SSIM manifest that our results can reserve abundant structural texture and edge details. This is because our SwinFuse has a strong feature extraction capacity with a pure transformer encoding backbone, and the global features may possess a more favorable representation ability than the local features. Moreover, the largest AG, SD and VIFF indicate that our fusion images acquire higher definition and contrast. On the one hand, our SwinFuse introduces a self-attention mechanism, and the extracted attentional maps focus more on the salient information of source images. On the other hand, we develop a feature fusion strategy based on $L_{1}$-norm, and make the fusion images well retain infrared competitive brightness information and visible distinct texture details. However, MI of our methods is a competitive value, the possible reason may be that the row and column vector normalization is proposed, and lead to a feature tradeoff in simultaneously retaining infrared thermal features and visible structural details. Even so, with multi-index evaluation, our SwinFuse achieves the best fusion performance.

\begin{table*}[!t]
	\renewcommand\arraystretch{1.5}
	\scriptsize
	%% increase table row spacing, adjust to taste
	%\renewcommand{\arraystretch}{1.3}
	% if using array.sty, it might be a good idea to tweak the value of
	% \extrarowheight as needed to properly center the text within the cells
	\caption{The computational efficiency comparisons of different methods on three datasets (Unit: second).}
	\label{table3}
	\centering
	%% Some packages, such as MDW tools, offer better commands for making tables
	%% than the plain LaTeX2e tabular which is used here.
	\begin{tabular}{ l c c c c c c c c c c}
		\hline
		Methods & MDLatLRR & IFCNN & DenseFuse & RFN-Nest & FusionGAN & GANMcC & PMGI & SEDRFuse & Res2Fusion & Ours \\
		\hline
		TNO &7.941 & 0.046 & 0.086 & 0.018 & 2.015  & 4.211  & 0.544 & 2.676 & 18.86 & 0.215\\
		Roadscene & 38.39 & 0.022 & 0.041 & 0.086 & 1.093 &  2.195 & 0.293 & 1.445 & 4.267 & 0.129 \\
		OTCBVS & 19.56 & 0.011 & 0.023 & 0.052 &  0.491 & 1.017 & 0.126 & 0.803 & 1.337 & 0.097\\
		\hline 
	\end{tabular}
\end{table*}

\subsection{Experiments on OTCBVS Dataset}

The public OTCBVS benchmark includes 12 video and image datasets, in which 31 image pairs selected from the OSU color-thermal database are used to demonstrate the generalization ability of our SwinFuse. Fig.13 shows a subjectively comparable example of \textit{video\_1036}. Compared with other methods, our SwinFuse obtains better intensity distribution with a clear target edge for the typical pedestrian target. Meanwhile, it also achieves more realistic scene detail for the parking lock. From all the above subjective comparisons, our SwinFuse has more superior fusion performance in maintaining infrared intensity distribution and visible texture details. Moreover, Fig.14 presents the objectively comparable results of the OTCBVS dataset, and our SwinFuse wins the first bank for AG, SF, SD, MS\_SSIM, SCD and VIFF, the second and third banks for MI and FMI\_w, respectively. In general, from the experimental results of three different datasets, the obtained optimal indexes of our SwinFuse are almost consistent, and its fusion performance is superior to other nine compared methods. 

In addition, we continue to test the computational efficiency of our SwinFuse. Notably, all the methods are tested on the GPU except for the traditional MDLatLRR, which is performed on the CPU. Table III gives their comparable efficiency. From these results, our computational efficiency follows behind IFCNN, DenseFuse and RFN-Nest, because their methods constructed an ordinary network architecture with several convolutional layers, and designed an average addition fusion strategy. However, our SwinFuse has a competitive computational efficiency, which is based on Swin Transformer hierarchical architecture, and has a linear computational complexity to image size. Therefore, we can conclude that our SwinFuse has better fusion performance, stronger generalization ability and competitive computational efficiency. Meanwhile, compared with the CNN, the pure transformer encoding backbone may be a more effective way to extract deep features for the fusion tasks.

\section{Conclusion}

In this paper, we present a residual Swin Transformer fusion network for infrared and visible images. our SwinFuse consists of three main parts: the global feature extraction, fusion layer and feature reconstruction. Especially, we build a fully attentional feature encoding backbone to model the long-range dependency, which only adopts a pure transformer without convolutional neural networks. The obtained global features have a stronger representation ability than the local features extracted by the convolutional operations. Moreover, we design a novel feature fusion strategy based on $L_{1}$-norm for sequence matrices, and measure the activity of source images from row and column vector dimensions, which can well retain competitive infrared brightness and distinct visible details.

We conduct a mass of experiments on the TNO, Roadscene and OTCBVS datasets, and compare it with other nine state-of-the-art traditional and deep learning methods. The experimental results demonstrate that our SwinFuse is a simple and strong fusion baseline, and achieves remarkable fusion performance with strong generalization ability and competitive computational efficiency, transcending other methods in subjective observations and objective comparisons. In future work, through overcoming the hand-craft fusion strategy, we will develop our SwinFuse to an end-to-end model, and extend it to settle other fusion tasks such as multi-focus and multi-exposure images.

% if have a single appendix:
%\appendix[Proof of the Zonklar Equations]
% or
%\appendix  % for no appendix heading
% do not use \section anymore after \appendix, only \section*
% is possibly needed

% use appendices with more than one appendix
% then use \section to start each appendix
% you must declare a \section before using any
% \subsection or using \label (\appendices by itself
% starts a section numbered zero.)
%

%\appendices
%\section{Proof of the First Zonklar Equation}
%Appendix one text goes here.

% you can choose not to have a title for an appendix
% if you want by leaving the argument blank
%\section{}
%Appendix two text goes here.

% use section* for acknowledgment
%\section*{Acknowledgment}
%This work is supported by the Applied Basic Research Project of Shanxi Province under Grant 201901D111260, the Open Foundation of Shanxi Key Laboratory of Signal Capturing \& Processing under Grant ISPT2020-4, the Startup Foundation for Doctors of Taiyuan University of Science and Technology under Grant 20162004.

% Can use something like this to put references on a page
% by themselves when using endfloat and the captionsoff option.
\ifCLASSOPTIONcaptionsoff
  \newpage
\fi

% trigger a \newpage just before the given reference
% number - used to balance the columns on the last page
% adjust value as needed - may need to be readjusted if
% the document is modified later
%\IEEEtriggeratref{8}
% The "triggered" command can be changed if desired:
%\IEEEtriggercmd{\enlargethispage{-5in}}

% references section

% can use a bibliography generated by BibTeX as a .bbl file
% BibTeX documentation can be easily obtained at:
% http://mirror.ctan.org/biblio/bibtex/contrib/doc/
% The IEEEtran BibTeX style support page is at:
% http://www.michaelshell.org/tex/ieeetran/bibtex/
%\bibliographystyle{IEEEtran}
% argument is your BibTeX string definitions and bibliography dataset(s)
%\bibliography{IEEEabrv,../bib/paper}

\begin{thebibliography}{1}
	
\bibitem{8765608}
Z.~Feng, J.~Lai and X.~Xie, ``Learning modality-specific representations for visible-infrared person re-identification,'' \emph{IEEE Trans. Image Process.}, vol.~29, pp.~579-590, 2020.	
	
\bibitem{ZHANG20201666}
X.~Zhang, P.~Ye, H.~Leung, K.~Gong and G.~Xiao, ``Object fusion tracking based on visible and infrared images: A comprehensive review,'' \emph{Inf. Fusion}, vol.~63, pp.~166-187, 2020.
	
\bibitem{9454273}
W.~Zhou, Y.~Zhu, J.~Lei, J.~Wan and L.~Yu, ``CCAFNet: Crossflow and cross-scale adaptive fusion network for detecting salient objects in RGB-D images,'' \emph{IEEE Trans. Multimedia}, 2021. doi: 10.1109/TMM.2021.3077767.


\bibitem{2015Multi}
Z.~Wang, F.~Yang, Z.~Peng, L.~Chen, and L.~Ji, ``Multi-sensor image enhanced fusion algorithm based on NSST and top-hat transformation,'' \emph{Optik}, vol. 126, no.~23, pp. 4184-4190, 2015.

\bibitem{9018389}
H.~Li, X.~Wu and J.~Kittler, ``MDLatLRR: A novel decomposition method for infrared and visible image fusion,'' \emph{IEEE Trans. Image Process.}, vol.~29, pp.~4733-4746, 2020.

\bibitem{0Infrared}
Z.~Wang, J.~Xu, X.~Jiang, and X.~Yan, ``Infrared and visible image fusion via hybrid decomposition of NSCT and morphological sequential toggle operator,'' \emph{Optik}, vol.~201, no.~1, 2020, Art no.~163497.

\bibitem{9146912}
Y.~Yang, Y.~Zhang, S.~Huang, Y.~Zuo and J.~Sun, ``Infrared and visible image fusion using visual saliency sparse representation and detail injection model,'' \emph{IEEE Trans. Instrum. Meas.}, vol.~ 70, pp.~1-15, 2021, Art no.~5001715.

\bibitem{2017Infrared}
J.~Ma, Z.~Zhou, B.~Wang, and H.~Zong, ``Infrared and visible image fusion based on visual saliency map and weighted least square optimization,'' \emph{Infr. Phys. Technol.}, vol.~82, pp. 8-17, 2017.

\bibitem{9626016}
Y.~Yang, W.~Zhang, S.~Huang, W.~Wan, J.~Liu and X.~Kong, ``Infrared and visible image fusion based on dual-kernel side window filtering and S-shaped curve transformation,'' \emph{IEEE Trans. Instrum. Meas.}, vol.~71, pp.~1-15, 2022, Art no.~5001915.

\bibitem{HU2019102977}
P.~Hu, F.~Yang, H.~Wei, L.~Ji and D.~Liu, ``A multi-algorithm block fusion method based on set-valued mapping for dual-modal infrared images,'' \emph{Infr. Phys. Technol.}, vol.~102, 2019, Art no.~102977.

\bibitem{ZHANG2021323}
H.~Zhang, H.~Xu, X.~Tian, J.~Jiang and J.~Ma, ``Image fusion meets deep learning: A survey and perspective,'' \emph{Inf. Fusion}, vol.~76, pp. 323-336, 2021.

\bibitem{8580578}
H.~Li and X.~Wu, ``Densefuse: A fusion approach to infrared and visible
images,'' \emph{IEEE Trans. Image Process.}, vol.~28, no.~5, pp.
2614-2623, 2019.

\bibitem{2019FusionGAN}
J.~Ma, W.~Yu, P.~Liang, C.~Li and J.~Jiang, ``Fusiongan: A generative
adversarial network for infrared and visible image fusion,''
\emph{Inf. Fusion}, vol.~48, pp.~11-26, 2019.

\bibitem{9710580}
Z.~Liu, Y.~Liu, Y.~Cao, H.~Hu, Y.~Wei, Z.~Zhang, S.~Lin and B.~Guo,  ``Swin Transformer: Hierarchical Vision Transformer using Shifted Windows,'' in \emph{Proc. IEEE/CVF Int. Conf. Comput. Vis. (ICCVW)}, 2021, pp.~9992-10002.

\bibitem{2014TNO}
A.~Toet(2014).\emph{TNO Image Fusion Dataset}. Figshare.Data.[Online]. 
Available: https://figshare.com/articles/TN\_Image\_Fusion\_Dataset/1008029.

\bibitem{2013A}
A.~Vaswani, N.~Shazeer, N.~Parmar, J.~Uszkoreit, L.~Jones, A.~N~Gomez, L.~Kaiser,
and I.~Polosukhin,  ``Attention is all you need,'' \emph{preprint arXiv:1706.03762}, 2017.

\bibitem{2020An}
A.~Dosovitskiy, L.~Beyer, A.~Kolesnikov, D.~Weissenborn, X.~Zhai, T.~Unterthiner, M.~Dehghani, M.~Minderer, G.~Heigold, S.~Gelly, J.~Uszkoreit and N.~Houlsby ``An image is worth 16x16 words: Transformers for image recognition at scale,'' in \emph{Proc. Int. Conf. Learn. Represent.(ICLR)}, 2021.

\bibitem{9658539}
H.~Dong, L.~Zhang and B.~Zou, ``Exploring vision transformers for polarimetric SAR image classification,'' \emph{IEEE Trans. Geosci. Remote Sens.}, vol.~60, pp.~1-15, 2022, Art no. 5219715.

\bibitem{9658539}
T.~Li, Z.~Zhang, L.~Pei and Y.~Gan, ``HashFormer: Vision Transformer based deep hashing for image retrieval,'' \emph{IEEE Signal Process. Lett.}, 2022. doi: 10.1109/LSP.2022.3157517.

\bibitem{9607618}
J.~Liang, J.~Cao, G.~Sun, K.~Zhang, L.~Van~Gool and R.~Timofte, ``SwinIR: Image Restoration Using Swin Transformer,'' in \emph{Proc. IEEE/CVF Int. Conf. Comput. Vis. (ICCVW)}, 2021, pp.~1833-1844.

\bibitem{lin2021swintrack}
L.~Lin, H.~Fan, Y.~Xu and H.~Ling,  ``SwinTrack: A Simple and Strong Baseline for Transformer Tracking,'' \emph{preprint arXiv:2112.00995}, 2021.

\bibitem{9345717}
H.~Xu, X.~Wang and J.~Ma, ``DRF: Disentangled representation for visible and infrared image fusion,'' \emph{IEEE Trans. Instrum. Meas.}, vol.~70, pp.~1-13, 2021, Art no.~5006713.

\bibitem{9416507}
J.~Ma, L.~Tang, M.~Xu, H.~Zhang and G.~Xiao, ``STDFusionNet: An infrared and visible image fusion network based on salient target detection,'' \emph{IEEE Trans. Instrum. Meas.}, vol.~70, pp.~1-13, 2021, Art no. 5009513.

\bibitem{2020IFCNN}
Y.~Zhang, Y.~Liu, P.~Sun, H.~Yan, X.~Zhao, and L.~Zhang, ``Ifcnn: A general
image fusion framework based on convolutional neural network,''
\emph{Inf. Fusion}, vol.~54, pp. 99-118, 2020.

\bibitem{2020SEDRFuse}
L.~Jiang, X.~Yang, Z.~Liu, G.~Jeon, M.~Gao and D.~Chisholm, ``SEDRFuse: A symmetric encoder-decoder with residual block network for infrared and visible image fusion,'' \emph{IEEE Trans. Instrum. Meas.}, vol.~70, pp.~1-15, 2021, Art no.~5002215.

\bibitem{9670874}
Z.~Wang, Y.~Wu, J.~Wang,J.~Xu and W.~Shao, ``Res2Fusion: Infrared and visible image fusion based on dense Res2net and double non-local attention models,'' \emph{IEEE Trans. Instrum. Meas.}, vol.~71, pp.~1-12, 2022, Art no.~5005012.

\bibitem{9528393}
Z.~Wang, J.~Wang, Y.~Wu, J.~Xu and X.~Zhang, ``UNFusion: A unified multi-scale densely connected network for infrared and visible image fusion,'' \emph{IEEE Trans. Circuits Syst. Video Technol.}, 2021. doi: 10.1109/TCSVT.2021.3109895.

\bibitem{LI202172}
H.~Li, X.~Wu and J.~Kittler, ``RFN-Nest: An end-to-end residual fusion network for infrared and visible images,'' \emph{Inf. Fusion}, vol.~73, pp.~72-86, 2021.

\bibitem{2020NestFuse}
H.~Li, X.~Wu and T.~Durrani, ``Nestfuse: An infrared and visible image
fusion architecture based on nest connection and spatial/channel attention models,'' \emph{IEEE Trans. Instrum. Meas.}, vol.~69, no.~12, pp.~9645-9656, 2020.

\bibitem{2020Rethinking}
H.~Zhang, H.~Xu, Y.~Xiao, X.~Guo and J.~Ma, ``Rethinking the image fusion: A fast unified image fusion network based on proportional maintenance of gradient and intensity,'' in \emph{Proc. AAAI Conf. Artif. Intell.}, vol.~34, no.~7, pp.~12797-12804, 2020.

\bibitem{2020U2Fusion}
H.~Xu, J.~Ma, J.~Jiang, X.~Guo and H.~Ling, ``U2fusion: A unified unsupervised image fusion network,'' \emph{IEEE Trans. Pattern Anal. Mach. Intell.}, vol.~44, no.~1, pp.~502-518, 2022.

\bibitem{9031751}
J.~Ma, H.~Xu, J.~Jiang, X.~Mei and X.~Zhang, ``DDcGAN: A dual-discriminator conditional generative adversarial network for multi-resolution image fusion,''
\emph{IEEE Trans. Image Process.}, vol.~29, pp.~4980-4995, 2020.

\bibitem{9274337}
J.~Ma, H.~Zhang, Z.~Shao, P.~Liang and H.~Xu, ``GANMcC: A generative adversarial network with multiclassification constraints for infrared and visible image fusion,'' \emph{IEEE Trans. Instrum. Meas.}, vol.~70, pp.~1-14, 2021. Art no.~5005014.

\bibitem{9520770}
Y.~Yang, J.~Liu, S.~Huang, W.~Wan, W.~Wen and J.~Guan,, ``Infrared and visible image fusion via texture conditional generative adversarial network,'' \emph{IEEE Trans. Circuits Syst. Video Technol.}, vol.~31, no.~12, pp.~4771-4783, 2021.

\bibitem{9216075}
J.~Li, H.~Huo, C.~Li, R.~Wang, C.~Sui and Z.~Liu, ``Multigrained attention network for infrared and visible image fusion,'' \emph{IEEE Trans. Instrum. Meas.}, vol.~70, pp.~1-12, 2021.

\bibitem{9670874}
J.~Li, H.~Huo, C.~Li, R.~Wang and Q.~Feng, ``AttentionFGAN: Infrared and visible image fusion using attention-based generative adversarial networks,'' \emph{IEEE Trans. Multimedia}, vol.~23, pp.~1383-1396, 2021.

\bibitem{2020Rethinking}
L.~Qu, S.~Liu, Y.~Xiao, M.~Wang and Z.~Song, ``TransMEF: A transformer-based multi-exposure image fusion framework using self-supervised multi-task learning,'' in \emph{Proc. AAAI Conf. Artif. Intell.}, 2022.

\bibitem{10.1007/978-3-319-10602-1_48}
T.~Lin, M.~Maire, S.~Belongie, J.~Hays, P.~Perona, D.~Ramanan,
P.~Doll{\'a}r and C.~L. Zitnick, ``Microsoft coco: Common objects in
context,'' in \emph{Computer Vision -- ECCV 2014}, D.~Fleet, T.~Pajdla, B.~Schiele, and T.~Tuytelaars, Eds.\hskip 1em plus 0.5em minus 0.4em\relax Cham: Springer International Publishing, 2014, pp. 740-755.

\bibitem{2020Roadscene}
H.~Xu(2020). \emph{Roadscene Database}. [Online]. Available: https://
github.com/hanna-xu/RoadScene.

\bibitem{2020Roadscene}
S.~Ariffin(2016). \emph{OTCBVS Database}. [Online]. Available: http://vcipl-okstate.org/pbvs/bench/.

\bibitem{2011Objective}
Z.~Liu, E.~Blasch, Z.~Xue, J.~Zhao, R.~Laganiere and W.~Wu, ``Objective assessment of multiresolution image fusion algorithms for context enhancement in night vision: A comparative study,'' \emph{IEEE Trans. Pattern	Anal. Mach. Intell.,} vol.~34, no.~1, pp. 94-109, 2011.

\bibitem{1997In}
Y.~Rao, ``In-fibre bragg grating sensors,'' \emph{Meas. Sci. Technol.}, vol.~8, no.~4, pp. 355-375, 1997.

\bibitem{7120119}
K.~Ma, K.~Zeng and Z.~Wang, ``Perceptual quality assessment for multi-exposure image fusion,'' \emph{IEEE Trans. Image Process.,} vol.~24, no.~11, pp.~3345-3356, 2015.

\bibitem{2002Information}
G.~Qu, D.~Zhang and P.~Yan, ``Information measure for performance of image fusion,'' \emph{Electron. Lett.,} vol.~38, no.~7, pp.~313-315, 2002.

\bibitem{477498}
A.~Eskicioglu and P.~Fisher, ``Image quality measures and their performance,'' \emph{IEEE Trans. Commun.}, vol.~43, no.~12, pp.~2959-2965, 1995.

\bibitem{V2015A}
V. Aslantas and E. Bendes, ``A new image quality metric for image fusion: The sum of the correlations of differences,'' \emph{AEU-Int. J. Electron. C.}, vol.~69, no.~12, pp. 1890-1896, 2015.

\bibitem{2013A}
Y.~Han, Y.~Cai, Y.~Cao, and X.~Xu, ``A new image fusion performance metric
based on visual information fidelity,'' \emph{Inf. Fusion}, vol.~14,
no.~2, pp. 127-135, 2013.











\end{thebibliography}
%
% <OR> manually copy in the resultant .bbl file
% set second argument of \begin to the number of references
% (used to reserve space for the reference number labels box)

% biography section
% 
% If you have an EPS/PDF photo (graphicx package needed) extra braces are
% needed around the contents of the optional argument to biography to prevent
% the LaTeX parser from getting confused when it sees the complicated
% \includegraphics command within an optional argument. (You could create
% your own custom macro containing the \includegraphics command to make things
% simpler here.)
%\begin{IEEEbiography}[{\includegraphics[width=1in,height=1.25in,clip,keepaspectratio]{mshell}}]{Michael Shell}
% or if you just want to reserve a space for a photo:

\begin{IEEEbiography}[{\includegraphics[width=1in,height=1.25in,clip,keepaspectratio]{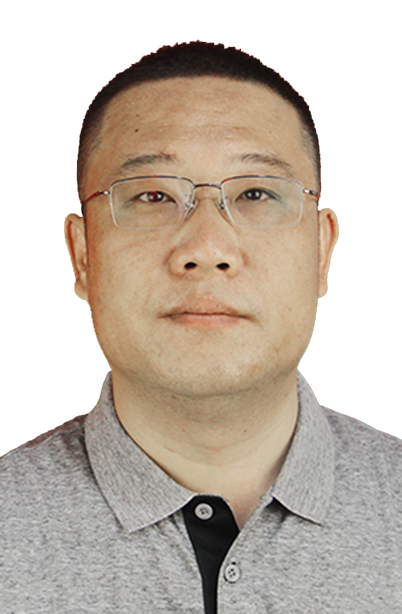}}]{Zhi-She Wang} \textit{(Member,~IEEE)} received the B.S degree in automation from North China Institute of Technology, Taiyuan, China, in 2002. He received the M.S. and Ph.D. degree in signal and information processing from the North University of China, Taiyuan, China, in 2007 and 2015. He is currently an associate professor with Taiyuan University of Science and Technology. His current research interests include computer vision, pattern recognition and machine learning.
\end{IEEEbiography}

% if you will not have a photo at all:
\begin{IEEEbiography}[{\includegraphics[width=1in,height=1.25in,clip,keepaspectratio]{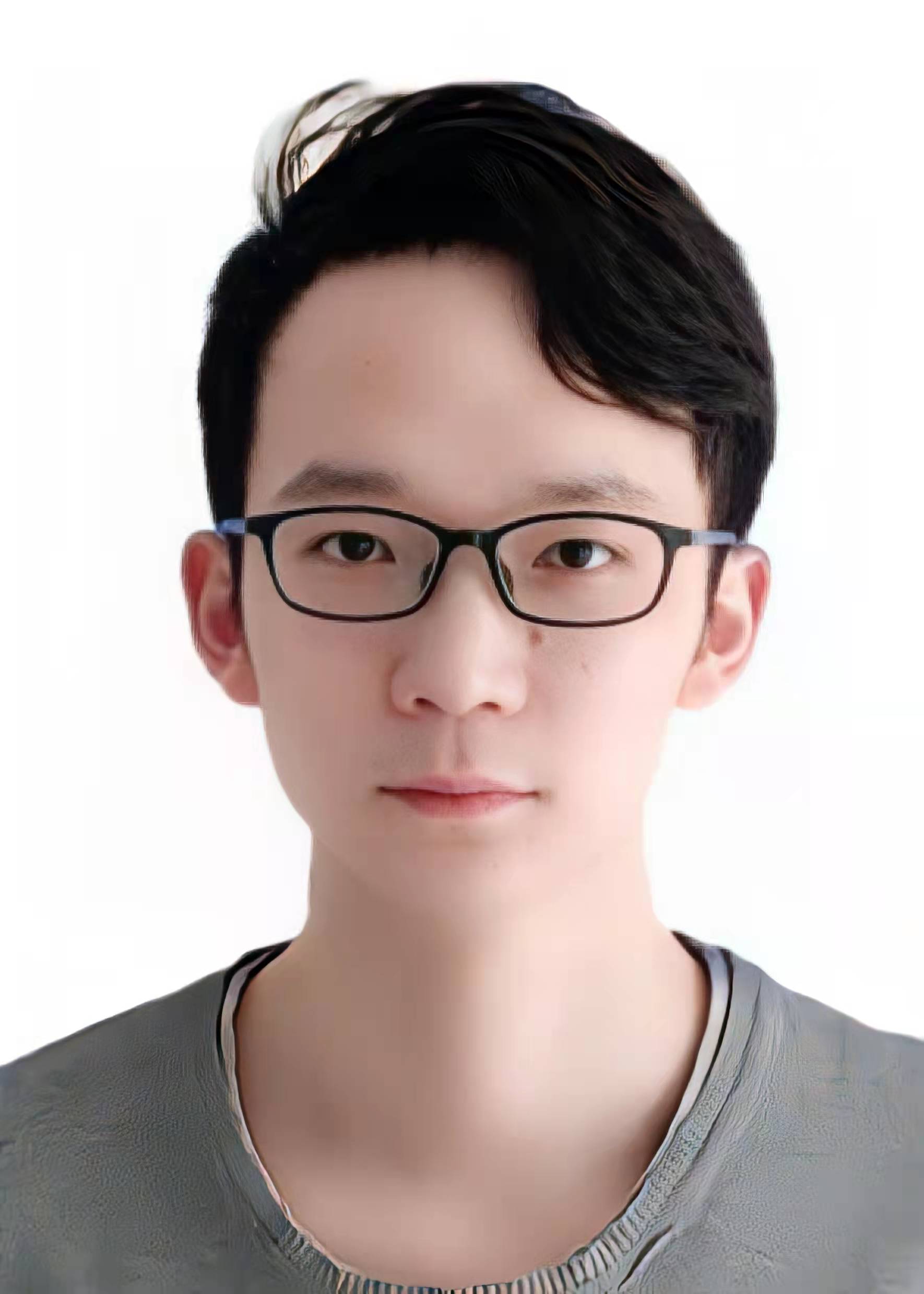}}]{Yan-Lin Chen}
	received the B.S degree in communication engineering from Hunan University of Technology, Zhuzhou, China, in 2019. He is currently pursuing the M.S. degree in electronic information at Taiyuan University of Science and Technology, Taiyuan, China. His current research interests include image fusion and deep learning.
\end{IEEEbiography}

\begin{IEEEbiography}[{\includegraphics[width=1in,height=1.25in,clip,keepaspectratio]{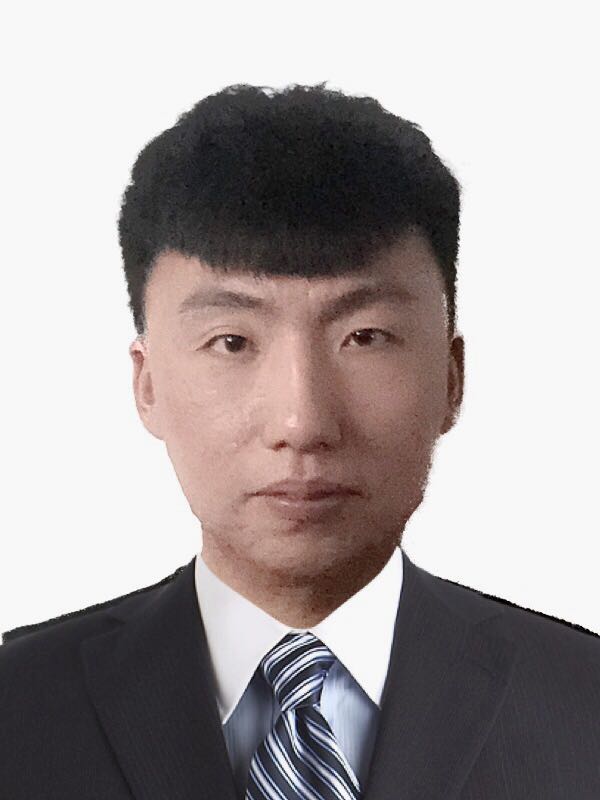}}]{Wen-Yu Shao}
	received the B.S degree in engineering mechanics from Taiyuan University of Science and Technology,Taiyuan, China, in 2020. He is currently pursuing the M.S. degree in electronic information at Taiyuan University of Science and Technology, Taiyuan, China. His current research interests include image fusion and deep learning.
\end{IEEEbiography}

\begin{IEEEbiography}[{\includegraphics[width=1in,height=1.25in,clip,keepaspectratio]{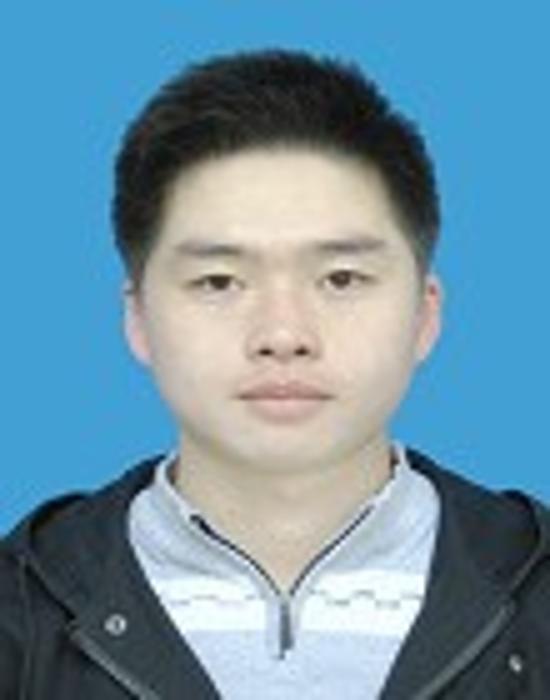}}]{Hui Li} received the B.Sc. degree in School of Internet of Things Engineering from Jiangnan University, China, in 2015. He received the PhD degree at the School of Internet of Things Engineering, Jiangnan University, Wuxi, China, in 2022. He is currently a Lecturer at the School of Artificial Intelligence and Computer Science, Jiangnan University, Wuxi, China. His research interests include image fusion and multi-modal visual information processing..
        
\end{IEEEbiography}

\begin{IEEEbiography}[{\includegraphics[width=1in,height=1.25in,clip,keepaspectratio]{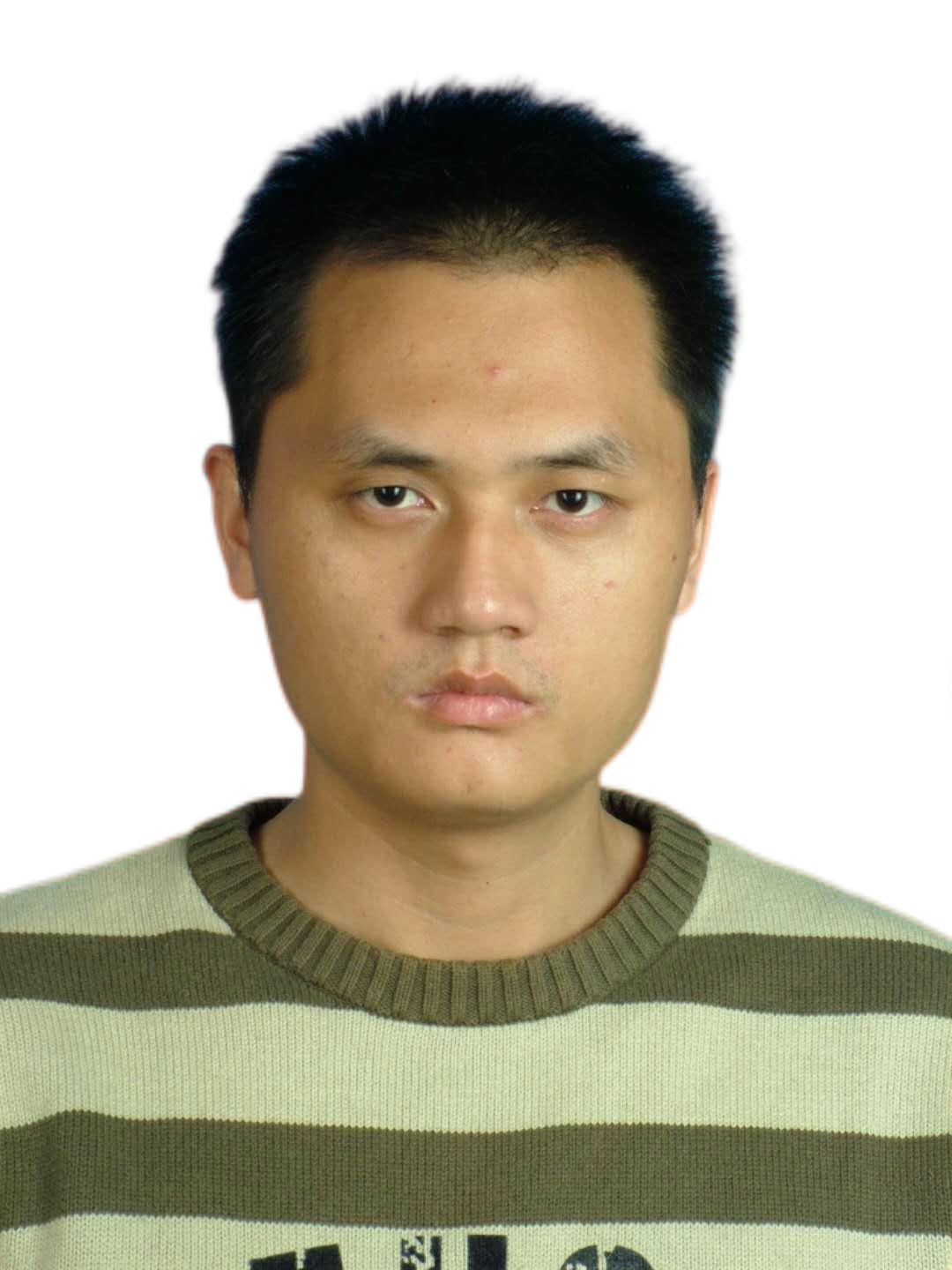}}]{Lei Zhang} received the B.S degree in electronic and information engineering from North University of China, Taiyuan, China, in 2004. He received the M.S and Ph.D.degree in information and communication engineering from the North University of China, Taiyuan, China, in 2007 and 2018. He is the currently an associate professor with Nanyang Normal University. His research interests include multi-modal image fusion, computer vision and machine learning.
\end{IEEEbiography}

% insert where needed to balance the two columns on the last page with
% biographies
%\newpage

% You can push biographies down or up by placing
% a \vfill before or after them. The appropriate
% use of \vfill depends on what kind of text is
% on the last page and whether or not the columns
% are being equalized.

%\vfill

% Can be used to pull up biographies so that the bottom of the last one
% is flush with the other column.
%\enlargethispage{-5in}

% that's all folks
\end{document}